\documentclass[11pt, a4paper, logo, copyright, nonumbering]{deepmind}

\usepackage[authoryear, sort&compress, round]{natbib}
\usepackage{caption}
\usepackage{soul}
\usepackage{subcaption}
\usepackage[ruled, lined, linesnumbered, commentsnumbered, longend]{algorithm2e}
\usepackage{amsbsy}
\usepackage{multirow}
\usepackage{diagbox}
\usepackage{adjustbox}
\usepackage{rotating,tabularx}

\title{Leveraging Jumpy Models for Planning and Fast Learning in Robotic Domains}

\correspondingauthor{zhangjingwei@deepmind.com}


\reportnumber{} 

\author[1]{Jingwei Zhang}
\author[1]{Jost Tobias Springenberg}
\author[1]{Arunkumar Byravan}
\author[1]{Leonard Hasenclever}
\author[1]{Abbas Abdolmaleki}
\author[1]{Dushyant Rao}
\author[1]{Nicolas Heess}
\author[1]{Martin Riedmiller}

\affil[1]{DeepMind}

\begin{abstract}
In this paper we study the problem of learning multi-step dynamics prediction models (jumpy models) from unlabeled experience and their utility for fast inference of (high-level) plans in downstream tasks. 
In particular we propose to learn a jumpy model alongside a skill embedding space offline, from previously collected experience for which no labels or reward annotations are required.
We then investigate several options of harnessing those learned components in combination 
with model-based planning or model-free reinforcement learning (RL) to speed up learning on downstream tasks.
We conduct a set of experiments in the RGB-stacking environment \cite{lee2022beyond}, 
showing that planning with the learned skills 
and the associated model
can enable zero-shot generalization to new tasks, and can further  speed up training of policies via reinforcement learning. These experiments demonstrate that jumpy models which incorporate temporal abstraction can facilitate planning in long-horizon tasks in which standard dynamics models fail.
\end{abstract}

\begin{document}
\maketitle

\section{Introduction}
\label{sec:intro}

From daily interactions with the world,
humans gradually develop an internal understanding of which series of events
would be triggered when a certain sequence of actions is taken 
\citep{maus2013motion,nortmann2015primary,hogendoorn2018predictive}.
This mental model of the world can serve as a compact proxy of our previous experiences and
help us plan out routes to desired goals before taking action
\citep{ha2018world}.
Studies have further implied that these mental predictive models might not be restricted to the level of primitive actions
\citep{botvinick2008hierarchical,consul2022improving},
but rather consider predictions over larger timescales
that abstract away detailed behavior consequences,
which can enable efficient long-horizon planning to guide our daily decision making.

When developing intelligent artificial agents it is therefore natural to imagine a similar process being useful for learning and transferring abstract models of the world across streams of experiences and tasks. We expect such a temporally abstract model of actions and dynamics to be significantly more useful than a simple one-step prediction model (together with primitive policies) when transferring them to a target task. This is because they should allow us to rapidly plan over long trajectories (to find some states with high rewards) while alleviating the common problem of error accumulation that occurs when chaining one-step prediction models which limits the effective planning horizon in most existing methods, e.g. \citep{byravan2020imagined,hafner2019dream,finn2017deep}.

When learning a temporally abstract model to predict the consequences of action sequences, 
ideally,
we might expect to train multi-step models that predict the (distribution of) 
ending states of arbitrary open-loop action sequences.
However,
the number of possible action sequences grows exponentially in the planning horizon,
while most such sequences lead to uninteresting behavior. At the same time the number of representative distinct action sequences that can be extracted from an offline dataset of interesting behaviour (i.e. a dataset stemming from executing explorative policies) is potentially much smaller. As a result a sensible approach then is to focus on learning jumpy models only
for those action sequences (or skills) which occur frequently in observed trajectories
and are therefore more likely to be generally useful and transferable across tasks
\cite{pmlr-v119-hasenclever20a,merel2018neural,merel2020catch,lynch2020learning,liu2022motor}.

In this paper we are interested in studying one approach that can be seen as a first step in this direction: learning multi-step (jumpy) dynamics prediction models together with temporally abstract skills. We will investigate this proposal by following a two-stage approach: Phase-1: offline learning of a jumpy model and multi-step skills based on (unlabeled) experience collected via some exploratory policies; followed by Phase-2: online reinforcement learning, or planning, utilizing the learned jumpy model and skills to speed up learning on a target task.
We note that this setting is different from offline policy learning
\citep{levine2020offline,wang2020critic,gulcehre2020rl} which requires offline data to contain transitions and rewards from agents attempting to solve the task at hand. We here assume no existence of rewards for the offline data, but only assume that the behaviour contained in the data set covers the parts of the state-space relevant for the downstream / target task. 

In this paper,
we thus propose to learn both a continuous skill embedding space (representing different action sequences) and a jumpy prediction model that predicts the consequences of choosing actions according to a skill.
With these components learned,
we study several ways of utilizing them to solve designated tasks downstream,
ranging from zero-shot transfer with model-based planning 
to learning from scratch with model-free reinforcement learning (RL).
We validate our approach in a set of experiments and show that 
the proposed method can enable zero-shot 
generalization to new tasks
with jumpy planning
and the learned skill embedding as well as that jumpy planning outperforms planning over primitive actions for the tasks we consider.
\section{Related work}

Previous works have looked into training action-conditioned one-step transition models
in various contexts.
In the setting of model-based reinforcement learning,
there is literature that use the learned models as differentiable environment simulators through which
the gradients of the expected return can directly pass 
\cite{heess2015learning,sanchez2018graph,hafner2019dream,byravan2020imagined,hafner2020mastering}.
This leads to low-variance gradients for updating the parameterized policy.
Another model-based direction
leverages conducting planning with learned models,
during which 
the predictive models are iteratively rolled out to generate imaginary trajectories.
This literature ranges from the early work of Dyna
\citep{sutton1991dyna} where planning is used to facilitate policy/value function learning
to methods that use learned models for decision-time planning
\cite{finn2017deep,ebert2018visual,sanchez2018graph,hafner2019learning,nagabandi2020deep}.
There also exist algorithms like MuZero
\citep{schrittwieser2020mastering}
that integrate planning in both training and inference time.
In 
\cite{byravan2021evaluating},
besides conducting model-predictive control (MPC) with a learned model,
they additionally study the influence of a learned action proposal policy for planning,
trained off the data generated with planning in the loop.
Their experiments on transfer and generalization show reasonable gains in data efficiency brought about by the learned model, 
which might be further improved with a bigger search budget.

Existing work in model-based approaches typically
learn one-step transition models and plan with a relatively short horizon
(e.g. less than 10 timesteps).
This is to reduce computational cost
as well as to avoid large compounding errors in planning.
This limits the use of learned one-step models for long-horizon tasks.
Therefore,
instead of learning predictive models for single step transitions conditioned on primitive actions,
learning multi-step models could possibly ameliorate this issue
\cite{liu2020hierarchical}.
Our work can be understood as taking a step in this direction where 
predictive models are trained to work on larger timescales,
thus could potentially avoid modelling high frequency noises and oscillations,
while focusing on longer lasing trends that model higher-level behavior patterns.

As discussed in the introduction,
in order to make it feasible to learn jumpy predictions,
we would need to target modelling multi-step transitions of 
only the most common action sequences in a dataset. I.e. we would have to condition our model on the types of action sequences/policies we are interested in, possibly by using a `latent' conditioning vector that summarizes the behavior in a given trajectory snippet. 
A number of approaches for extracting temporally abstract skills exist in the literature.
Neural probabilistic motor primitives (NPMP)
\citep{merel2018neural}
compresses an expert dataset of humanoid control
by encoding trajectory snippets into a continuous latent space of motor intentions and learn a skill conditioned policy by decoding actions from pairs of skills and states.
To encourage temporally correlated latent skills to be nearby in the latent embedding space,
an auto-correlated prior is enforced.
Since NPMP does not train a model of the learned latent skills,
an RL agent is trained over the skill space 
in order to perform new tasks.
Play-supervised latent motor plans (Play-LMP)
\citep{lynch2020learning}
also learns a continuous skill embedding space from human teleoperated play data.
They follow a slightly different setup from NPMP in that
their skill policy is goal-conditioned.
Additionally,
they learn a plan recognition module that encodes trajectories into latent skills and a plan proposal module that encodes pairs of starting and goal states into skills
and at the same time minimizes the KL-divergence between the plan recognition and the plan proposal.
Similarly, there exists work on extracting discrete options from offline data (representing multi-task data with a `mixture of skills' approach) \citep{wulfmeier2021data} as well as approaches that combine discrete and continuous abstractions \citep{rao2021learning}; since skill models are not learned during skill extraction,
they
do not directly support zero-shot generalization/transfer with hierarchical planning.
For skill extraction approaches that also consider skill model learning,
\cite{salter2022model} builds on \cite{wulfmeier2021data} and additionally learns option models to encourage the extraction of predicable options.
While the extracted options could be beneficial for planning,
they do not investigate in this regard.
The follow-up work to \cite{lynch2020learning},
broadly-exploring local-policy trees (BELT)
\citep{sermanet2021broadly}
is most relevant to our work.
They build on the skill embedding from Play-LMP and 
additionally learn skill-conditioned prediction models to be used in 
a rapid-exploring random tree (RRT) style planning procedure.  
In contrast to this, our approach learns the skill embedding and its corresponding model in parallel (rather than separating out model and skill learning), additionally we do not follow a goal-conditioned formulation and rely on less involved planning procedures.

The aforementioned methods focus on learning reusable skills form offline data.
In the online (reinforcement) learning setting,
the agent has to to learn skills via active interactions with the environment, which brings with it the exploration problem of discovering skills in the first place.
Under multi-task/goal scenarios
\cite{teh2017distral},
\cite{galashov2019information,goyal2018transfer,tirumala2020behavior}
impose a form of information asymmetry between agent policy and an additionally learned default policy,
such that the latter is enforced to represent reusable behaviors across related tasks,
without access to goal/task-specific information.
These methods generally encourage exploration by injecting entropy regularization to the optimization objective
or when latent variables are used to represent the behavioral context
\cite{goyal2018transfer},
adding exploration bonus using the KL-divergence 
between the encoder that encodes the latent from state and goal versus one that encodes from the state only.
There are also methods like
\cite{heess2016learning,peng2017deeploco}
that learn two level control hierarchies that besides information asymmetry,
also impose control rate asymmetry to the different levels of controllers. 
In the case of unsupervised skill discovery in the absence of rewards/tasks,
there is a line of research basing on empowerment maximization
\cite{salge2014empowerment}. 
These methods train agents by maximizing the intrinsic reward of empowerment,
which is the mutual information (MI) between a sequence of actions and the final state reached,
conditioned on a starting state.
Discovering skills via maximizing empowerment means 
finding the set of skills with which the number of reliably reachable distinct final states is maximized.
Variational information maximisation 
\citep{mohamed2015variational}
proposes a scalable approach to approximate empowerment with variational inference and function approximation,
but their approach operates with open-loop action sequences which can be very restrictive and can lead to underestimation of empowerment
\citep{gregor2016variational}.
Variational intrinsic control (VIC)
\citep{gregor2016variational}
instead learns closed-loop option/skill-conditioned policies and show improved performance.
Dynamics-aware discovery of skills (DADS)
\citep{sharma2020dynamics}
differs from previous methods as it uses the model-based version of MI
in which it learns the skill dynamics,
predicting the distribution of next state given the current state and a skill.
This skill conditioned one-step transition model enables planning to be conducted for new tasks.
Contrastive intrinsic control (CIC) \citep{laskin2021cic}
proposes a new estimator for MI that improves on generating diverse behaviors as well as discriminating different skills.
In order to separate out the concern of exploration,
in this work we aim to extract skills from offline datasets before using them for a target task.

\section{Methods}
\label{sec:methods}

We consider learning from offline data.
In particular,
the offline dataset $\mathcal{D}$ contains 
$M$ trajectory snippets of the form
$\tau_m=\{s_t, a_t, \cdots, s_{t+K-1}, a_{t+K-1}, s_{t+K}\}$
with states 
$s\in\mathcal{S}$ and
actions $a\in\mathcal{A}$,
where $K$ controls the context length.
All trajectories are collected within the same environment 
with initial state distribution
$p(s_0)$
and transition dynamics 
$p(s_{t+1} \mid s_t, a_t)$,
with each trajectory being collected by executing a random sequence of pre-specified basic exploration policies\footnote{
Two exploration policies are randomly selected per episode, each running for 200 steps,
more details in experiments.
}.

Our method then operates in two phases: 
1) a learning phase in which we learn a jumpy model and skill embedding from the offline dataset; and 2)
a harnessing phase where the extracted components 
act as compact representations of the offline dataset
to solve designated tasks.

\phantomsection
\subsection{Phase-1 - Learning: Extracting Low-level Policy and Jumpy Model}
\label{sec:phase-1}

\begin{figure}[t]
    \centering
    \includegraphics[width=0.8\textwidth]{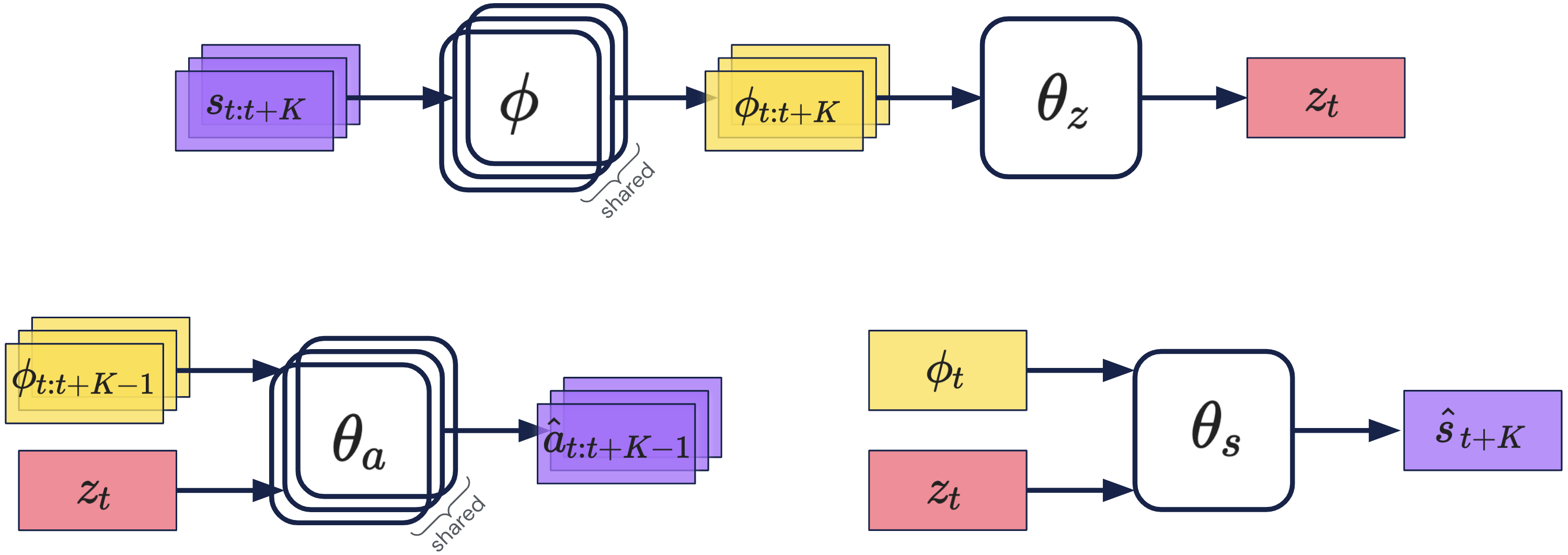}
    \captionsetup{font=small}
    \caption{
    Schematic plot of the proposed architecture, 
    with the top row showing the encoder and the bottom the action decoder (low-level policy) and the jumpy state decoder (jumpy model).
    }
    \label{fig:jumpy}
\end{figure}

This phase aims to learn a skill embedding space along with a jumpy model from the offline dataset,
for which we propose a VAE-style architecture \citep{kingma2013auto} composed of an encoder and two decoders
(Figure \ref{fig:jumpy}).

As shown in the top row of Figure \ref{fig:jumpy},
the proposed approach takes as input the states contained in
trajectory snippets of length $K+1$:
$s_{t:{t+K}}=\{s_t, s_{t+1}, \cdots, s_{t+K}\}$.
These states are first mapped through the same set of parameters $\phi$ of a fully connected network into state features
$\phi_{t:t+K}$
(shorthanded for $\{\phi(s_t),\cdots,\phi(s_{t+K})\}$);
applying the same transformation to all states within a snippet
means that the extracted features are expected to only capture information that are indicative of inferring the high level trend within the snippet,
but not necessarily those that are specific to each corresponding timestep.
These $K+1$ state features are then encoded together into a continuous latent code
$z_t \sim q(z_t \mid \phi_{t:{t+K}}; \theta_{z})$
by an encoder parameterized by $\theta_{z}$.

The decoder network consists of an action decoder 
(parameterized by $\theta_a$) 
and a jumpy state decoder 
(parameterized by $\theta_s$). 
Concatenating each state feature
with the latent code $z_t$,
the action decoder learns to predict each corresponding
immediate action
by
maximizing the likelihood
$\prod_{k=0}^{K-1}p(a_{t+k} \mid \phi_{t+k},z_t; \theta_a)$.
The jumpy state decoder takes in $z_t$ and the feature of the first state within the snippet $\phi_t$ and from those predicts the state that is $K$ steps ahead:
$p(s_{t+K} \mid \phi_t,z_t; \theta_s)$.
Jointly learning to decode the jumpy state and the low-level actions from the latent code $z_t$
ensures that the latent 
encodes relevant information for predicting both;
if only the skill-conditioned action decoder
is learned during training
while its corresponding model is learned afterwards,
or vice versa,
it is likely that the information contained
in the learned latent encoding would not be sufficient,
as inferring the immediate action
and inferring the jumpy state
should likely require different sets of cues.

The latent $z$ space can be interpreted from two perspectives,
on the one hand it encodes skills
while on the other it encodes effects.
Firstly,
the resulting action decoder could be deployed as a latent-conditioned low-level (LL) policy,
therefore the latent $z$'s can be interpreted as 
skills or high-level (HL) actions 
\citep{merel2018neural,lynch2020learning}
that encode different behavior modes;
in turn, the jumpy state decoder
could be viewed as predicting the state that would be reached in $K$ steps 
when following a certain latent skill.
Alternatively,
if we first consider the jumpy model that uses $z$ to decode jumpy states, 
then the latent $z$ space could be understood as the embedding space that encodes the effect \citep{whitney2019dynamics} or change 
that could be brought about to the state
with a $K$-step horizon;
in this case the action decoder could be seen as the policy 
that would lead to the desired effect or change indicated by $z$.
These two interpretations are complementary in the sense that 
the former one is looking at the cause while the latter one focuses on the effect. 

The overall loss for learning the encoder and the two decoders is
\begin{align}
    \mathcal{L}(\phi,\theta_{z},&\theta_{a},\theta_{s})    
=
    \mathop{\mathbb{E}}_{\{s_t,a_t,\cdots,s_{t+K}\} \sim \mathcal{D}}
    \Big[
        D_{\text{KL}}
        \big(
            q(z_t \mid \phi_{t:t+K}; \phi,\theta_z) \parallel \mathcal{N}(0,I)
        \big)
\label{eq:loss-kl}
    \\&-
        \beta_{a} \sum_{k=0}^{K-1} \log p(a_{t+k} \mid \phi_{t+k},z_t; \phi,\theta_a)
    -
        \beta_{s} p(s_{t+K} \mid \phi_t,z_t; \phi,\theta_s)
    \Big],
\label{eq:loss-recon}
\end{align}
which amounts to
minimizing the KL-divergence between the variational posterior and a standard normal prior
while maximizing the log-likelihood of the action and jumpy state predictions
($I$ denotes the identity matrix,
$\beta_a$, $\beta_s$ are loss coefficients).
We note that in all experiments we assume isotropic normal distributions for both decoders 
so minimizing the negative-log-likelihood terms in Eq.\eqref{eq:loss-recon} reduces to minimizing L2 losses.

One important detail to note is that
since the ranges of raw state values vary much in scale,
we find it helpful to interpret the state decoder predictions 
as state deltas $\hat{s}_{t+K}-s_t$ 
and normalized within $[-1,1]$,
which are then scaled back to the original range 
and added onto $s_t$ to recover the predicted jumpy states.

\subsection{Phase-2 - Harnessing: Utilizing Learned Components}
\label{sec:phase-2}

\begin{table}[b]
    \centering
    \small
    \begin{tabular}{c|ccc}
                                & Propose HL & Refine HL with Planning & LL \\ 
    \hline
    Random-HL                   & $\mathcal{N}(0,I)$ & & $p(a_t \mid s_t, z_t)$ \\
    Zero-shot via Planning      & $\mathcal{N}(0,I)$ & $p(s_{t+K} \mid s_t, z_t)$ & $p(a_t \mid s_t, z_t)$ \\
    RL-HL                       & $\pmb{\pi(z_t \mid s_t)}$         & & $p(a_t \mid s_t, z_t)$ \\
    RL-HL + Planning            & $\pmb{\pi(z_t \mid s_t)}$         & $p(s_{t+K} \mid s_t, z_t)$ & $p(a_t \mid s_t, z_t)$ \\
    RL-HL + Planning + Finetune & $\pmb{\pi(z_t \mid s_t)}$         & $\widetilde{p}(s_{t+K} \mid s_t, z_t)$ & $p(a_t \mid s_t, z_t)$ \\
    RL-from-scratch             &                 & & $\pmb{\pi(a_t \mid s_t)}$                              
    \end{tabular}
    \captionsetup{font=small}
    \caption{
    Different ways of harnessing learned components from phase-1 in phase-2. 
    Methods are categorized by how the high-level (HL) skills are proposed, 
    whether the proposed HL skill is refined with model-based planning facilitated by the learned jumpy model,
    and how the low-level (LL) action is selected.
    Methods ranging from zero-shot to learning from scratch are listed from top to bottom.
    Components that need to be learned from scratch in phase-2 are marked in bold,
    those that need to be finetuned are marked with tilde,
    all other components are either learned in phase-1 and can be directly deployed 
    or do not require any learning.
    }
    \label{tab:harnessing}
\end{table}

Once the initial learning phase is completed,
the resulting components,
namely the encoder that learns a continuous embedding space of high-level (HL) skills,
as well as the skill-conditioned low-level (LL) policy and its jumpy model,
can be harnessed to fulfill a given task. 
There exist multiple options to accomplish a designated task ranging from zero shot transfer (by planning with the learned model and executing the low-level skills for a new reward) to model-free or model-based RL to derive a new policy for the target task. We can differentiate the possible options by varying: 1) how we obtain the latent embedding $z$ used by the dynamics and skill modules, 2) whether we use the learned low-level (LL) action module or learn a new policy from scratch, 3) whether we use the learned dynamics model to plan for a better $z$ for a given state $s$ at action selection time. We consider multiple variations along these axes which are summarized in Table \ref{tab:harnessing}. We name each variation according to their interpretation:

The first option, "\textbf{Random-HL}", is not a practical option but listed for ease of discussion.
This is a method that only makes use of the learned LL policy but not the learned jumpy model.
Here a HL skill $z_t$ is simply sampled from the standard normal prior.
Then, conditioning on the sampled $z_t$,
a LL action can be selected using the learned skill-conditioned LL policy to be executed in the environment.
Since task information is not incorporated in any of the procedures of "Random-HL",
it is not expected to be able to accomplish given tasks. Instead we expect it to reproduce snippets of behaviour present in the dataset we used for training in phase-1.

\begin{algorithm}[t]
    \small
    \SetKwFunction{evalTraj}{evalTraj}
    \SetKwInOut{Parameter}{Parameter}

    \KwIn{
    $s_t$: current state;
    $p_0(z \mid s)$: initial skill proposal policy;
    $p(s_{t+K} \mid s_t, z_t)$: the learned jumpy model;
    $Score(\cdot)$: trajectory scoring function.
    }
    \KwOut{
    $z_t$: the next latent skill to follow.
    }
    \Parameter{
    $I$: number of CEM iterations;
    $N$: number of samples;
    $H$: planning horizon;
    $\eta$: elite fraction.
    }

    \For{$i \leftarrow 0$ \KwTo $I-1$}{
        $plans = [\ ]$\\
        $scores = [\ ]$\\
        \For{$n \leftarrow 0$ \KwTo $N-1$}{
            $\hat{s}_t=s_t$\\
            $plan_n = [\ ]$\\
            $trajectory_n = [\hat{s}_t]$\\
            \For{$h \leftarrow 0$ \KwTo $H-1$}{
                $\hat{z}_{t+h \cdot K} \sim p_i(\cdot \mid \hat{s}_{t+h \cdot K})$ \tcp*[f]{Sample latent skill from proposal.}\\
                $\hat{s}_{t+(h+1) \cdot K} \sim p(\cdot \mid \hat{s}_{t+h \cdot K}, \hat{z}_{t+h \cdot K})$ \tcp*[f]{Rollout jumpy model.}\\
                $plan_n.\text{append}(\hat{z}_{t+h \cdot K})$\\
                $trajectory_n.\text{append}(\hat{s}_{t+(h+1) \cdot K})$\\
            }
            $plans.\text{append}(plan_n)$\\
            $scores.\text{append}(Score(trajectory_n))$
        }
        Rank $plans$ by their $\text{scores}$ and retain the top $\eta \cdot N$.\\ 
        Update skill proposal policy to $p_{i+1}(z \mid s)$ using the statistics of the top $\eta \cdot N$ plans.
    }
    \KwRet{First latent skill of the highest scored plan in $plans$}

    \caption{CEM-MPC with Jumpy Model}
    \label{alg:planning}
\end{algorithm}

For "\textbf{Zero-shot via Planning}",
the proposals for $z_t$ are also sampled from the standard normal prior.
But before feeding the sampled $z$'s as conditioning to the LL policy,
the proposals are first refined for the given task at hand (as specified by a reward function $R(s_t) \in \mathbb{R}$) via model-based planning using the pretraind jumpy model.
We use model-predictive control (MPC) with cross-entropy methods (CEM) and outline the planning procedure in Algorithm \ref{alg:planning}.
Within each CEM iteration during planning,
$N$ samples of imaginary trajectories will be rolled out from the current state $s_t$. 
Each of the sampled trajectories involves $H$ iterative predictions of the next jumpy state 
from the previously predicted one using the jumpy model
and conditioning on the proposed latent skill,
with the final predicted state $H\cdot K$ timesteps ahead of $s_t$. 
Each of those $N$ rollouts will be evaluated according to a scoring function based on the task reward:
\begin{align}
    Score(\{s_t,s_{t+K},\cdots,s_{t+H\cdot K}\})=\sum_{h=0}^{H} \gamma_K^h R(s_{t+h\cdot K}),
    \label{eq:plan-score}
\end{align}
where $\gamma_K$ is the discount factor 
that operates on $K$-step intervals 
and $R(\cdot)$ is the reward function that defines the designated task. 
The top ranked samples will be used to update the skill proposal policy,
and the best sample resulting from the last CEM-iteration will be returned.
Then the first latent skill $z$ contained in this best plan will be used
in the learned LL policy to obtain an action $a_t \sim p(\cdot \mid s_t, z_t)$.
\footnote{
The appendix (Table \ref{tab:app-zero-shot}) contains additional results where we vary the planning horizon $H$ and the number of steps the selected $z$ is kept fixed for executing LL actions.
}
We note that this "Zero-shot via Planning" procedure is a zero-shot method for accomplishing a given task assuming knowledge of the reward function.
Here the HL skill is proposed from the standard normal prior,
and the planning only utilizes the jumpy model
and the LL policy which are both learned in phase-1 from offline data.
We also note that throughout our experiments,
the planning procedure contains only 1 CEM-iteration which essentially reduces to random shooting. 

"\textbf{RL-HL}"
is similar to "Random-HL" in that it
also does not involve planning. However, instead of directly using the prior as the HL proposal,
it learns a skill proposal policy using model-free RL
(here we use DMPO \cite{abdolmaleki2018maximum})
with the skill embedding $z$-space as its action space.
This gives a RL-learned high level policy $\pi(z_t \mid s_t)$,
which together with the low-level policy $p(a_t \mid s_t, z_t)$ learned during phase-1, can output primitive actions.
This is a common setup for reusing pretrained low-level skills in a model-free setting \cite{merel2018neural,liu2020hierarchical}.

"\textbf{RL-HL + Planning}" is similar to "Zero-shot via Planning"
as it also uses the planning procedure outlined in Algorithm \ref{alg:planning} to refine proposed HL skills
before feeding them to 
the LL policy.
But instead of getting the proposals from the prior,
it learns a proposal $\pi(z_t \mid s_t)$ the same way as in "RL-HL".
We note that in this case the data filling the replay buffer used for RL training
in phase-2
is generated by the planning procedure,
which means that the HL policy is trained off-policy using the data collected from executing the planning procedure (this use of a planner  as a policy is similar to \citep{springenberg2020local,byravan2021evaluating}). 

Another variation from this setup,
"\textbf{RL-HL + Planning + Finetune}"
additionally finetunes the jumpy model parameters 
($\theta_s$) alongside learning a HL proposal policy,
again using the data collected by executing MPC plans derived with the full planning procedure.
The motivation for this is that when the jumpy model is trained in phase-1,
it has only seen data distribution from the offline dataset,
which might greatly differ from the state distribution induced by the planner on the target task.
Therefore further finetuning the jumpy model on the planned data could potentially make the model and the deployed policy more consistent.

Lastly,
"\textbf{RL-from-scratch}" makes use of neither the learned jumpy model nor the LL policy
but learns a policy $\pi(a_t \mid s_t)$ from scratch with model-free RL 
(DMPO) on the primitive action space.

\section{Experiments}

\begin{table}[b]
    \centering
    \small
    \begin{tabular}{c|c }
         Base Policy & Behavior \\
         \hline
         open\_fingers & Open the gripper \\
         close\_fingers & Close the gripper \\
         reach\_red & Move the gripper to the red object \\
         lift\_red & Lift the red object off the ground \\
         red\_hover\_blue & Lift the red object and hover over the blue object \\
         move\_pinch\_x\_inc & Move the gripper in the positive x-axis direction \\
         move\_pinch\_y\_inc & Move the gripper in the positive y-axis direction \\
         move\_pinch\_z\_inc & Move the gripper in the positive z-axis direction \\
         move\_pinch\_x\_dec & Move the gripper in the negative x-axis direction \\
         move\_pinch\_y\_dec & Move the gripper in the negative y-axis direction \\
         move\_pinch\_z\_dec & Move the gripper in the negative z-axis direction \\
    \end{tabular}
    \captionsetup{font=small}
    \caption{The set of base policies used to generate training data for phase-1. All policies are trained via RL (DMPO \citep{abdolmaleki2018maximum,hoffman2020acme}) with shaped rewards encouraging the desired behaviour.}
    \label{tab:policies}
\end{table}

\begin{figure}[ht]
    \centering
    \footnotesize
    \begin{subfigure}{0.63\textwidth}
        \centering
        \includegraphics[width=\linewidth]{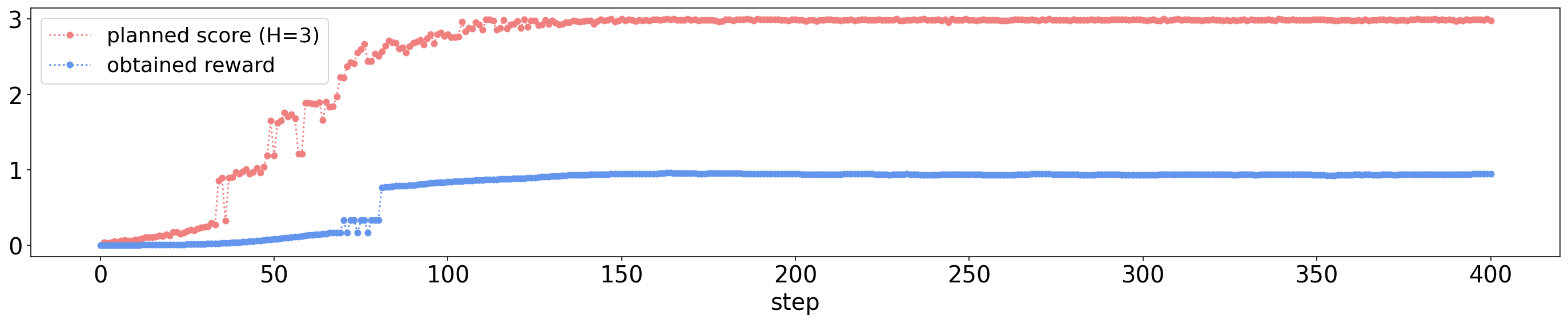}
        \captionsetup{font=footnotesize,justification=centering}
        \caption{
        Planned score VS. obtained reward during an evaluation episode for the task of red\_hover\_green.
        }
        \label{fig:vis-plot}
    \end{subfigure}
    \begin{subfigure}{.59\textwidth}
        \centering
        \includegraphics[width=\linewidth]{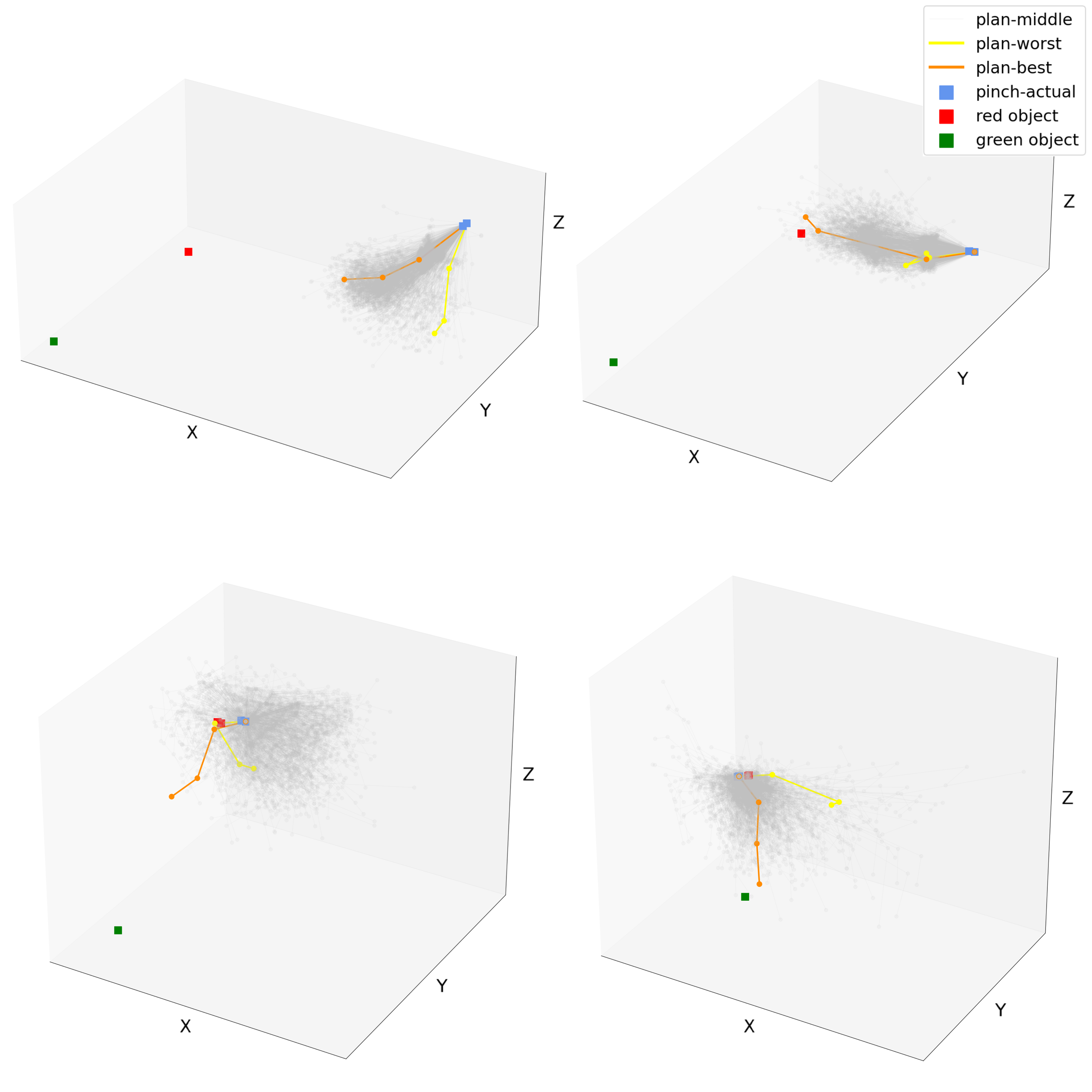}
        \captionsetup{font=footnotesize,justification=centering}
        \caption{
        Planning procedure visualizations of step 3, 44, 100, 151.
        }
        \label{fig:vis-trajs}
    \end{subfigure}
    \begin{subfigure}{.203\textwidth}
        \centering
        \includegraphics[width=\linewidth]{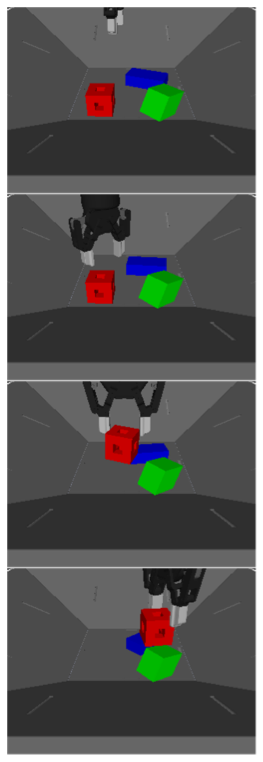}
        \captionsetup{font=footnotesize,justification=centering}
        \caption{Corresponding images.}
        \label{fig:vis-imgs}
    \end{subfigure}

    \captionsetup{font=footnotesize}
    \caption{
    Visualizations of the "Zero-shot via Planning" procedure for a red\_hover\_green episode.
    Fig.~\ref{fig:vis-plot} plots the planned score (Eq.~\ref{eq:plan-score}, $\gamma_K=1$) against the actual reward obtained per step,
    with a planning horizon $H=3$ (corresponds to $30$ steps in terms of primitive actions with $K=10$). 
    Fig.~\ref{fig:vis-trajs} shows the planning procedure for several exemplar steps
    (left to right, top to bottom)
    (corresponding images shown in Fig.~\ref{fig:vis-imgs}).
    The red and green cube shows the positions for the red and green object respectively.
    The sampled plans ($N=1000$) are shown as orange (best plan), yellow (worst plan) and grey (other plans) lines
    (plotted in terms of the pinch pose).
    The blue cubes show the actual poses of the pinch.
    }
    \label{fig:vis}
\end{figure}

We conducted a set of experiments with data collected from the RGB-stacking environment \cite{lee2022beyond} 
with a Sawyer robot arm simulator and a triplet of objects of color red, green and blue.
To collect the offline dataset $\mathcal{D}$,
we first select several 'base' policies for data collection that were trained with simple hand coded rewards (moving the arm along x, y or z axis in 3D-space, opening/closing the fingers), or have been previously trained to perform a certain more complex movement (i.e. reaching one of the objects, lifting them, or hovering one object over another object). The complete set of base policies we use for data collection is given in Table \ref{tab:policies}. We collect data by selecting 2 base policies at random per episode, each of them being executed for 200 timesteps, one after the other. All episodes collected this way make up our offline dataset $\mathcal{D}$\footnote{We note that for practical reasons, instead of storing a large dataset, we generate data on the fly into a small buffer for training in phase-1.} for phase-1.

The states and actions in the environment are continuous,
with the state being 86-dimensional 
and the action 5-dimensional. 
The context length $K$ is set to 10.
The state feature embedder
(parameterized by $\phi$)
contains three linear layers
of size $(256, 256, 128)$,
with a LayerNorm 
\citep{ba2016layer}
applied after the first linear layer
and an ELU activation after each layer.
This gives state features of size $128$.
The encoder 
(parameterized by $\theta_z$)
first aggregates
all the state features 
$\phi_{t:t+K}$
within one trajectory snippet
with a $1\times1$ convolution over the time dimension followed by ELU.
Then three linear layers of size
$(256, 256, 64)$
with ELU activation are applied.
This outputs the mean and standard deviation (each of size $32$) 
of the diagonal gaussian of the latent skill,
from which $32$-dimensional $z$'s can be sampled. 
The inputs to the action decoder are $K$ $160$-dimensional vectors,
concatenating each state feature $\phi_{t:{t+K-1}}$ with $z$,
while the jumpy state decoder takes in the concatenation of $\phi_t$ and $z$.
Both decoders contain four linear layers of size $256$ followed by ELU)
with final linear outputs of size $86$ and $5$ respectively.

After the learning phase is completed,
the harnessing phase is carried out in which
all variations listed in Table \ref{tab:harnessing} are tested.
We first conduct the evaluation for 
"Random-HL"
and "Zero-shot via Planning" as these two methods do not involve any further learning. 
To further clarify the planning procedure,
we visualize an exemplar evaluation episode of "Zero-shot via Planning" for the task red\_hover\_green in Fig.~\ref{fig:vis}.

The red\_hover\_green task expects the gripper to grasp the red object and hover it over the top of the green object (note that no such behaviour was present in the training data for phase-1).
For the episode shown in Fig.~\ref{fig:vis},
the planning is configured with the following parameters:
planning horizon $H=3$,
context length $K=10$,
number of samples $N=1000$
and discount factor $\gamma_K=1$ for the scoring function (Eq.~\ref{eq:plan-score}).
This means that each of the 1000 sampled plans involves 
a sequence of 3 applications of the jumpy model,
which corresponds to 30 steps in terms of primitive actions since each jumpy step predicts the jumpy state that is 10 steps ahead.
Those sampled plans are shown in Fig.~\ref{fig:vis-trajs} as orange (highest scored plans),
yellow (lowest scored plans)
and grey (all other plans) lines.
The best plan (orange) will be selected,
then a primitive action can be obtained using the first latent skill of that best plan.
The actual reward obtained by executing the primitive action is plotted against the planned score in Fig.~\ref{fig:vis-plot}.
Since the planned score is calculated on 3 iteratively predicted jumpy states,
it roughly has a scale 3 times of the immediate reward obtained.
From the visualised episode,
the best plans points towards the red object at the beginning
(first two figures in \ref{fig:vis-trajs}) since the gripper needs to first hold the red object in hand;
it then directs the gripper towards the green object once the red object has been grasped and lifted;
finally it plans to hover above the green object.

The evaluation results for zero-shot methods over a set of tasks are presented in Table \ref{tab:zero-shot}.
Besides the zero-shot approaches listed in Table \ref{tab:harnessing}:
"Random-HL" (presented as an indicator of the difficulty of each task)
and "Zero-shot via Planning"
(procedure visualized in Fig.~\ref{fig:vis} with context length $K=10$),
we additionally include the results of
"Zero-shot via Planning (not jumpy $K=1$)"
as an additional baseline (in which the model is learned on the primitive action scale and thus planning occurs in the original problem space).
For the tasks whose desired behaviors are contained in the training dataset of phase-1,
the scores obtained by the corresponding base policies are also listed.
Each configuration is evaluated with 50 episodes with random seeds,
with a maximum number of 400 steps for each evaluated episode.

For the in-distribution tasks
(reach\_red, lift\_red, red\_hover\_blue),
the jumpy planning ("Zero-shot via Planning") agent performs very well.
We emphasize that although the agent has been trained on trajectory snippets of these three tasks
during phase-1,
it was not trained explicitly to solve these control tasks but was just trained to 
model the trajectory data derived from executing policies for these tasks, thus this model seems to exhibit generalization capabilities.
It can also be observed that the non-jumpy planner ($K=1$) performs on par with the jumpy ($K=10$) version in the simplest in-distribution task reach\_red and on the simpler reach\_green and lift\_green out of distribution tasks, we hypothesize that this is due to the fact that long horizon planning is not critical for completing this simple task (and relatively greedy reward maximization can be competitive). However, it can be observed that the non-jumpy $K=1$ configuration performs poorly on most of the out-of-distribution tasks. In contrast, the jumpy planning agent ("Zero-shot via Planning") is able to solve all out-of-distribution tasks except the lift\_green task (which could be due to the fact that none of the episodes in the training data in phase-1 contains trajectories where the gripper has the green object in hand,
therefore the imaginary rollouts planned by the jumpy model could hardly lead to states where this task could be accomplished). This clearly demonstrates the utility of jumpy-planning.
Overall our results validate that the learned components from phase-1,
namely the latent skill embedding space,
the latent conditioned LL policy
and its jumpy model,
could facilitate the "Zero-shot via Planning" agent to solve these tasks zero-shot without the need for additional finetuning/learning.
The appendix contains further evaluation results (Table \ref{tab:app-zero-shot},\ref{tab:app-zero-shot-1step}).

\begin{table}[b]

\centering
\footnotesize

\begin{tabular}{c|ccc|c}
                  & Random-HL        & \begin{tabular}[c]{@{}c@{}}Zero-shot via Planning \\ (jumpy) \end{tabular} & \begin{tabular}[c]{@{}c@{}}Zero-shot via Planning\\ (not jumpy, K=1)\end{tabular} & Pretrained Base Policy \\ \hline
reach\_red        & $70.91\pm124.06$ & $373.78\pm27.99$       & $\mathbf{375.96\pm17.86}$                                                       & $378.43\pm11.64$       \\
lift\_red         & $15.63\pm27.28$  & $\mathbf{307.52\pm65.97}$       & $130.16\pm52.55$                                                       & $293.91\pm31.76$       \\
red\_hover\_blue  & $13.90\pm20.53$  & $\mathbf{278.66\pm60.88}$       & $73.41\pm20.02$                                                        & $321.85\pm48.75$       \\ \hline
red\_stack\_blue  & $8.25\pm12.20$   & $\mathbf{169.98\pm61.86}$       & $57.28\pm33.95$                                                        &                        \\
red\_hover\_green & $13.08\pm20.00$  & $\mathbf{266.28\pm79.53}$       & $84.78\pm40.51$                                                        &                        \\
red\_stack\_green & $7.80\pm12.98$   & $\mathbf{162.75\pm73.69}$       & $46.09\pm13.43$                                                        &                        \\
bring\_red        & $15.26\pm27.09$  & $\mathbf{225.97\pm83.76}$      & $132.52\pm100.68$                                                      &                        \\
reach\_green      & $36.01\pm51.38$  & $241.41\pm104.34$      & $\mathbf{375.54\pm16.06}$                                                       &                        \\
lift\_green       & $12.95\pm15.30$  & $59.75\pm26.74$        & $\mathbf{114.77\pm46.45}$                                                       &                       
\end{tabular}

\captionsetup{font=small}
\caption{
Evaluation results for zero-shot harnessing methods.
The table shows for each method 
the mean and standard deviation of the undiscounted cumulative reward
obtained within an episode
(the maximum number of steps for each episode is $400$)
over $50$ random seeds.
The highest return obtained in each task is marked in bold (excluding those obtained via the pretrained base policies).
Further evaluation results are presented in the appendix (Table \ref{tab:app-zero-shot},\ref{tab:app-zero-shot-1step}).
}
\label{tab:zero-shot}
\end{table}

\begin{figure*}[t]
    \small
    \captionsetup{font=small,justification=centering}
    \centering
    \begin{subfigure}{0.83\textwidth}
        \centering
        \includegraphics[width=\linewidth]{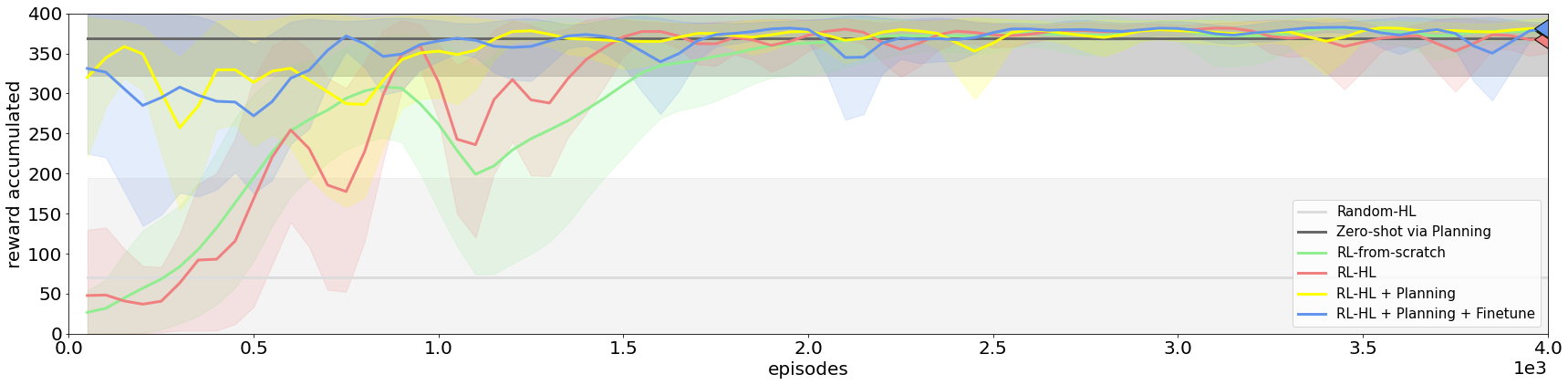}
        \caption{Accumulated reward on the reach\_red task.}
        \label{fig:reachred}
    \end{subfigure}
    ~ 
    \begin{subfigure}{0.83\textwidth}
        \centering
        \includegraphics[width=\linewidth]{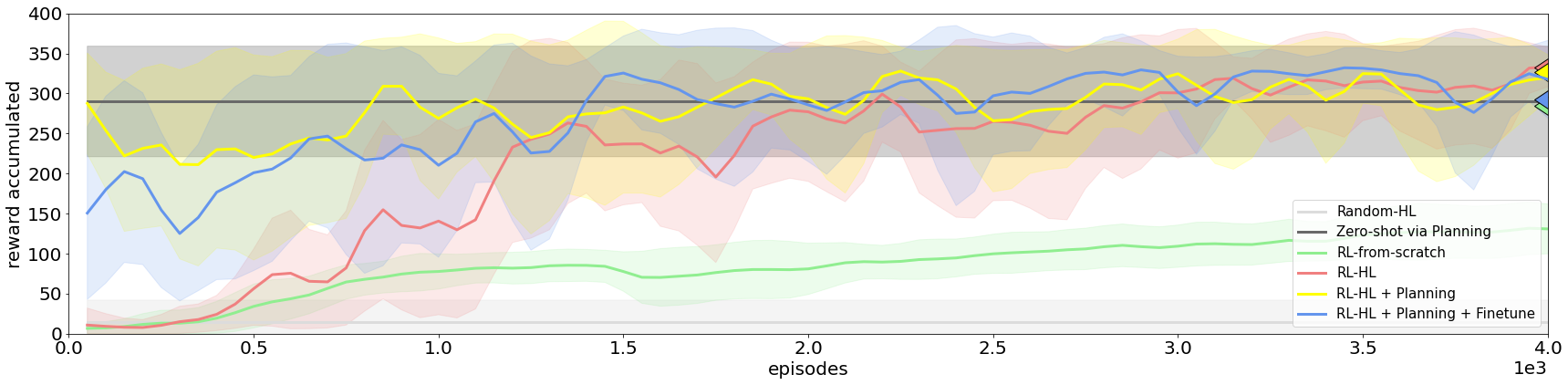}
        \caption{Accumulated reward on the lift\_red task.}
        \label{fig:liftred}
    \end{subfigure}
    ~ 
    \begin{subfigure}{0.83\textwidth}
        \centering
        \includegraphics[width=\linewidth]{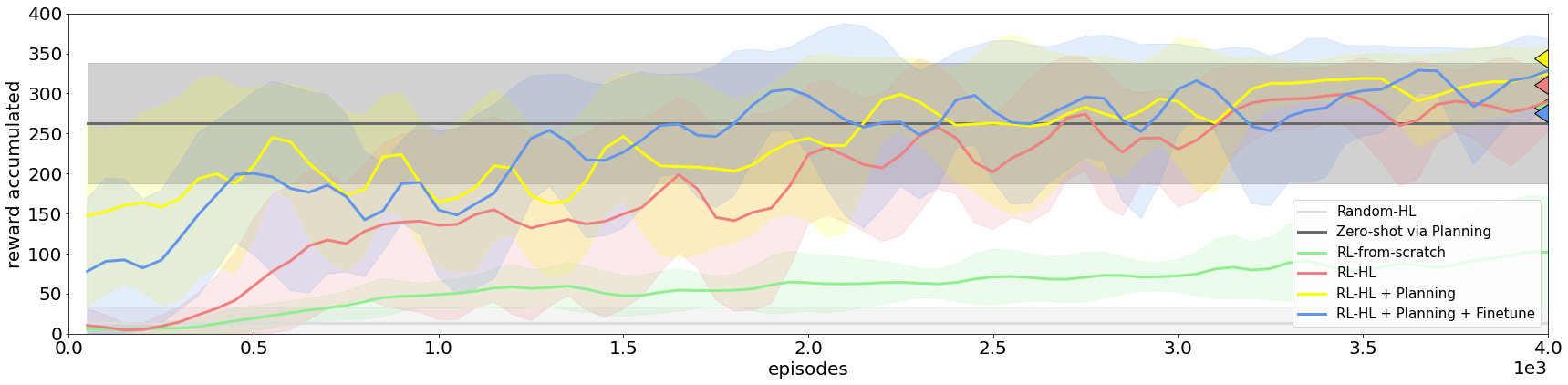}
        \caption{Accumulated reward on the red\_hover\_blue task.}
        \label{fig:redhoverblue}
    \end{subfigure}
    \captionsetup{font=small,justification=raggedright}
    \caption{
        Comparison of harnessing options on in-distribution tasks. 
        Each plot shows the mean and standard deviation of 5 random seeds 
        ("Random-HL" and "Zero-shot via Planning" plot the results from Table \ref{tab:zero-shot}, 
        each conducted with 50 random seeds),
        with left-pointing triangles indicating returns converged at step
        $4\mathrm{e}{3}$ (Fig.\ref{fig:reachred}),
        $3\mathrm{e}{4}$ (Fig.\ref{fig:liftred}),
        $1\mathrm{e}{4}$ (Fig.\ref{fig:redhoverblue})).
    }
    \label{fig:indist}
\end{figure*}

\begin{figure*}[p]
    \small
    \captionsetup{font=scriptsize,justification=centering}
    \centering
    \begin{subfigure}{0.78\textwidth}
        \centering
        \includegraphics[width=\linewidth]{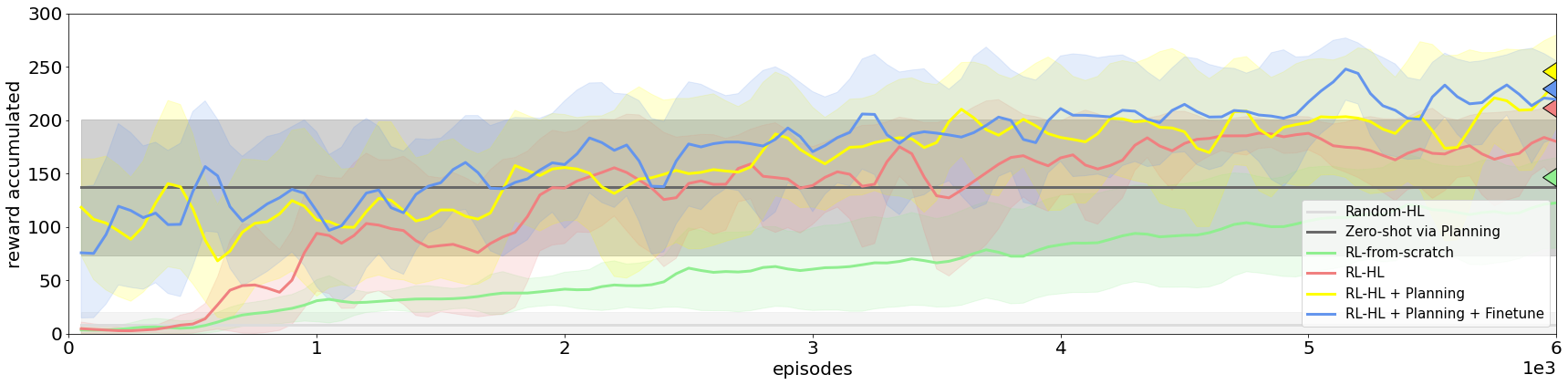}
        \caption{Accumulated reward on the red\_stack\_blue task.}
        \label{fig:redstackblue}
    \end{subfigure}
    ~ 
    \begin{subfigure}{0.78\textwidth}
        \centering
        \includegraphics[width=\linewidth]{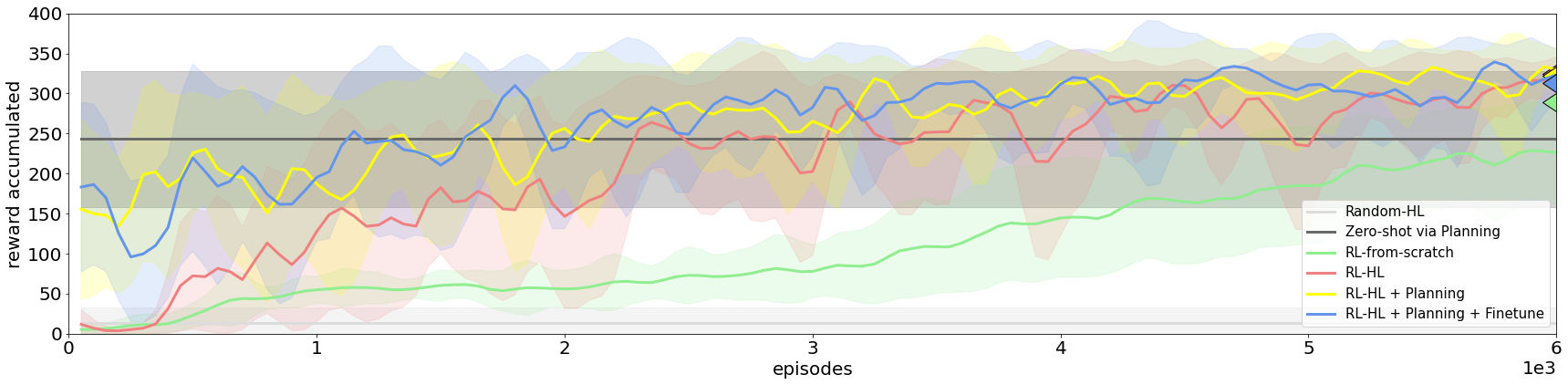}
        \caption{Accumulated reward on the red\_hover\_green task.}
        \label{fig:redhovergreen}
    \end{subfigure}
    ~ 
    \begin{subfigure}{0.78\textwidth}
        \centering
        \includegraphics[width=\linewidth]{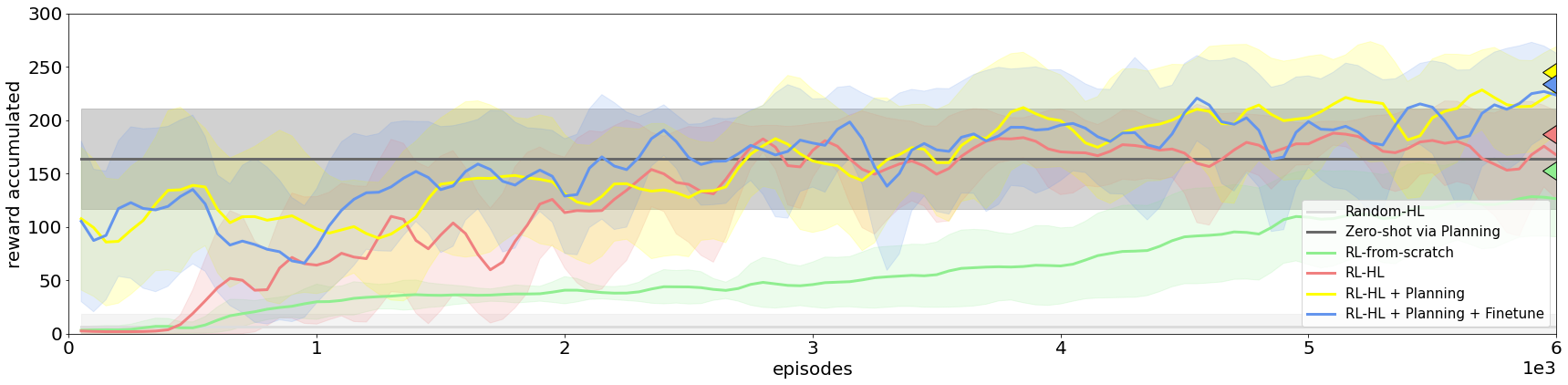}
        \caption{Accumulated reward on the red\_stack\_green task.}
        \label{fig:redstackgreen}
    \end{subfigure}
    ~ 
    \begin{subfigure}{0.78\textwidth}
        \centering
        \includegraphics[width=\linewidth]{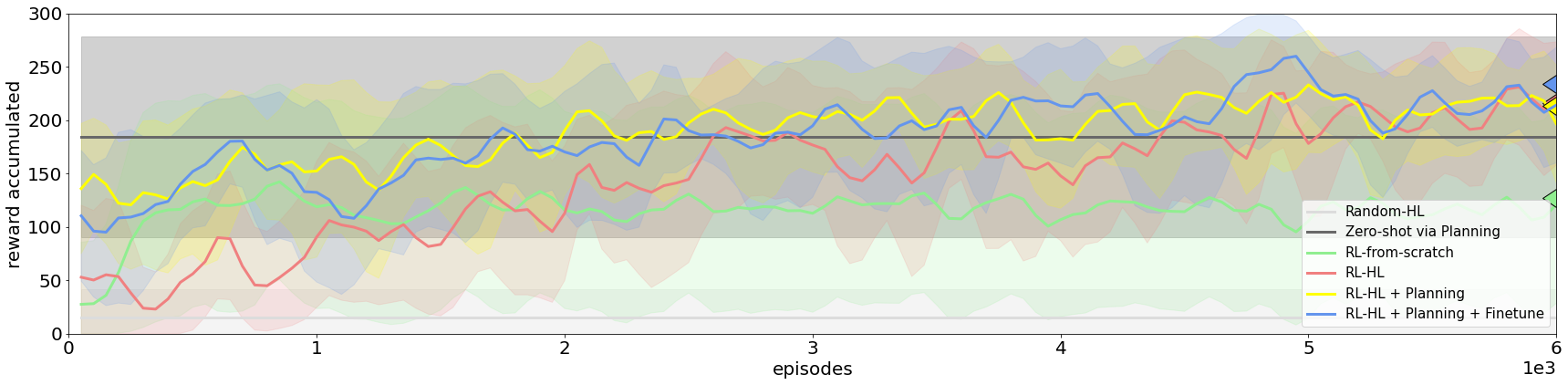}
        \caption{Accumulated reward on the bring\_red task.}
        \label{fig:bringred}
    \end{subfigure}
    ~ 
    \begin{subfigure}{0.78\textwidth}
        \centering
        \includegraphics[width=\linewidth]{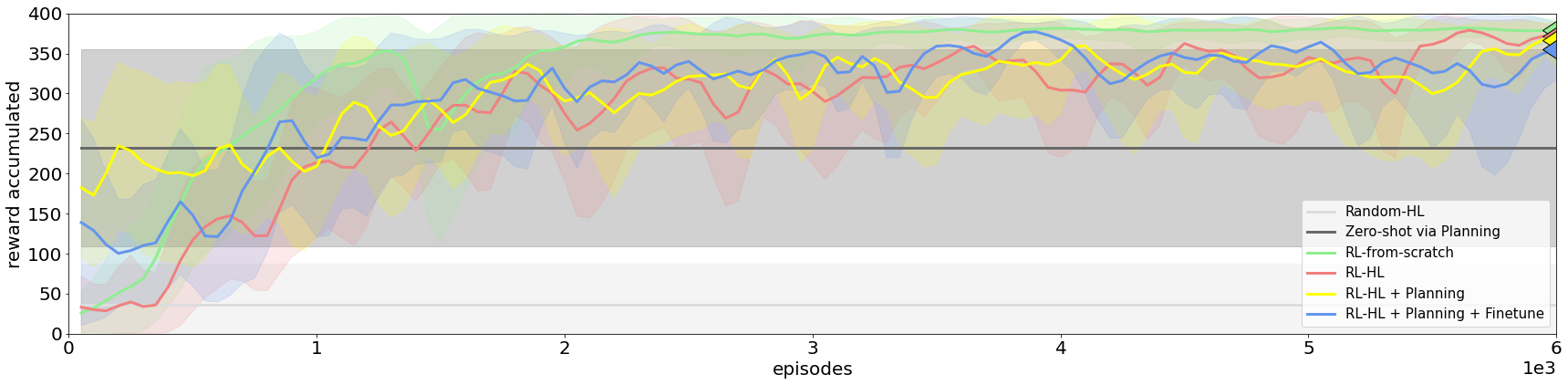}
        \caption{Accumulated reward on the reach\_green task.}
        \label{fig:reachgreen}
    \end{subfigure}
    ~ 
    \begin{subfigure}{0.78\textwidth}
        \centering
        \includegraphics[width=\linewidth]{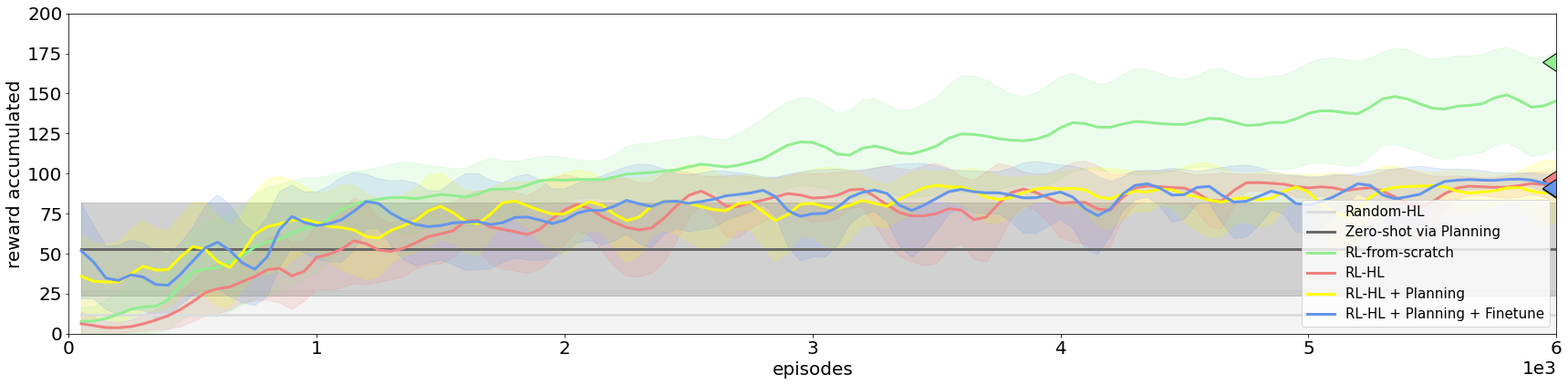}
        \caption{Accumulated reward on the lift\_green task.}
        \label{fig:liftgreen}
    \end{subfigure}
    \captionsetup{font=footnotesize,justification=raggedright}
    \caption{
        Comparison of harnessing options on out-of-distribution tasks. 
        Each plot shows the mean and standard deviation of 5 random seeds 
        ("Random-HL" and "Zero-shot via Planning" plot the results from Table \ref{tab:zero-shot}, 
        each conducted with 50 random seeds)
        with left-pointing triangles indicating returns converged at step $1\mathrm{e}{4}$.
    }
    \label{fig:ood}
\end{figure*}

Next,
we set out to evaluate other harnessing options which involve learning to potentially improve the transfer performance.
For each of the methods listed in Table \ref{tab:zero-shot} that requires additional learning,
we run experiments with five random seeds and plot their mean and standard deviation.
For ease of comparison,
the results of "Random-HL" and "Zero-shot via Planning" from Table \ref{tab:zero-shot} are also plotted.
The results for in distribution tasks are shown in Figure \ref{fig:indist}
and the out of distribution tasks are shown in Figure \ref{fig:ood}.
In general,
"Zero-shot via Planning"
gives strong performance for in-distribution tasks,
and its performance for out-of-distribution tasks are also descent.
This is promising as ideally we would prefer to be able to transfer previously extracted knowledge to new tasks zero-shot.
The learning for "RL-HL" takes off faster than "RL-from-scratch",
which shows that the learned skill embedding space provides a behavior abstraction 
that enables more efficient exploration than directly learning on the primitive action space.
"RL-HL + Planning" also learns efficiently as "RL-HL",
while having the extra boost in performance brought by planning at the beginning of training.
Additionally finetuning the jumpy model however does not seem to improve "RL-HL + Planning + Finetune" over "RL-HL + Planning",
possibly because the finetune data which are generated from planning still lie well 
in the distribution as the original training data,
since the planning procedure is carried out off the jumpy model itself.

\clearpage

Task wise,
we can see that the agent perform quite similarly on 
red\_hover\_blue (Fig.\ref{fig:redhoverblue}) and red\_hover\_green (Fig.\ref{fig:redhovergreen}),
although it has only been shown episodes for the former.
This could possibly be due to the fact that during training the blue object has appeared in various locations 
and therefore the jumpy model or the latent space could well cover the state space 
such that the planning procedure or the HL policy being learned could easily pick up.
This also shows that the learned components could enable reasonable zero-shot generalization
and boost few-shot transfer.
This is also shown in 
the bring\_red (Fig.\ref{fig:bringred}) task 
where it takes "RL-from-scratch" a long time to catch up with the performance of the other options.
However no effective transfer is shown on the lift\_green task (Fig.\ref{fig:liftgreen}).
As discussed for the "Zero-shot via Planning" experiment,
the agent has never seen any data with the gripper having the green object in hand.
In this case,
the abstraction handicaps the agent from learning useful policies since it is well out of distribution.

Overall,
through these conducted experiments,
it can be observed that the components learned in phase-1 of the proposed approach can act effectively as a compact representation of the training data via temporal abstraction.
This leads to reasonable zero-shot generalization (via planning) to tasks whose desired behaviors are at least partially contained in the training data,
with the planning process itself also being largely sped up due to the jumpy predictions.
The learned jumpy model and skill embedding also enable various more efficient few-shot adaptation options in phase-2 to further improve performance.

\section{Conclusions and Future Work}

To conclude,
we propose an algorithm to learn 
a skill embedding space as well as a jumpy model of the skills
from offline datasets of interactions.
These learned components could be harnessed in various ways to solve newly encountered tasks.
We have conducted experiments in the RGB-stacking environment
with a range of different transfer setups and
validated the efficacy of our proposed method, showing the benefits of jumpy-planning in learned latent skill spaces.

There are many potential future directions to study.
For example,
we could look into higher performance architectures such as transformers 
\cite{vaswani2017attention,brown2020language}
for the jumpy model.
We could also investigate how the distributions of the pretrained base policies 
for generating the training data could affect the generalization of the extracted skill embedding and skill models.
Also,
how to learn a skill space that is the most effective is generally an open research topic,
any improvements in this regard should potentially translate to better performance for our proposed method.
Additionally,
this paper has only looked at learning jumpy models of fixed look-ahead intervals 
(e.g., 10 time-steps ahead),
while it could potentially be more efficient to learn jumpy models with varying horizons,
adjusting its resolution according to the varying predictability/controllability
\cite{jayaraman2018time}
across different parts of the state space.
Another aspect would be
to investigate other training or finetuning options to bring cycle consistency of the model and the policy.
Furthermore,
we could iteratively build up a hierarchy of jumpy models that progressively abstracts the world model,
with which zero-shot generalization with planning for solving complicated real-world problems might be possible.

\section{Acknowledgements}
The authors would like to thank Tim Hertweck, Giulia Vezzani, Markus Wulfmeier, Thomas Lampe, Will Whitney, Jackie Kay, Sarah Bechtle, Francesco Nori, Andrea Huber, the Control Team, and many others at DeepMind for their helpful feedback, discussions, and support for this paper.

\clearpage
\newpage
\bibliography{main}

\clearpage
\newpage
\section{Appendix}

\begin{sidewaystable}
\scriptsize
\centering

\begin{tabular}{cc|cccccc}
                                                                              & 
\diagbox[width=100pt]{Horizon $H$\footnote{Latent planning horizon in $z$.}}{No. actions\footnote{Number of LL actions executed per planning iteration, with the planned latent kept fixed.}}
                                                                              & 1                 & 2                 & 10                & 20                & 100               & 200               \\ \hline
\multirow{4}{*}{reach\_red}                                                   & 1 (10 env steps)                                                           & $368.18\pm52.90$  & $\mathbf{373.78\pm27.99}$  & $368.13\pm46.15$  & $367.88\pm50.16$  & $292.28\pm172.73$ & $296.87\pm119.52$ \\
                                                                              & 2 (20 env steps)                                                        & $357.01\pm87.19$  & $367.01\pm57.49$  & $326.72\pm127.75$ & $321.27\pm128.51$ & $300.50\pm114.03$ & $320.62\pm105.77$ \\
                                                                              & 3 (30 env steps)                                                         & $342.47\pm104.61$ & $341.67\pm108.02$ & $343.73\pm94.87$  & $341.56\pm100.86$ & $271.35\pm140.34$ & $227.09\pm158.06$ \\
                                                                              & 5 (50 env steps)                                                          & $302.95\pm138.23$ & $336.56\pm116.26$ & $321.06\pm122.22$ & $316.33\pm129.47$ & $249.79\pm149.52$ & $240.14\pm160.09$ \\ \hline
\multirow{4}{*}{lift\_red}                                                    & 1 (10 env steps)                                                           & $280.12\pm84.81$  & $\mathbf{307.52\pm65.97}$  & $295.96\pm87.21$  & $244.82\pm104.21$ & $104.67\pm77.69$  & $91.80\pm52.26$   \\
                                                                              & 2 (20 env steps)                                                           & $267.14\pm121.12$ & $289.66\pm89.05$  & $283.85\pm94.13$  & $234.81\pm114.18$ & $117.46\pm81.03$  & $82.07\pm53.98$   \\
                                                                              & 3 (30 env steps)                                                           & $253.65\pm125.90$ & $291.76\pm103.41$ & $256.61\pm119.24$ & $230.29\pm124.29$ & $106.33\pm68.03$  & $80.25\pm48.49$   \\
                                                                              & 5 (50 env steps)                                                           & $218.59\pm134.35$ & $261.98\pm108.21$ & $247.16\pm118.90$ & $188.65\pm122.26$ & $98.72\pm79.70$   & $64.97\pm51.47$   \\ \hline
\multirow{4}{*}{red\_hover\_blue}  & 1 (10 env steps)                                                           & $227.79\pm91.05$  & $276.56\pm63.36$  & $\mathbf{278.66\pm60.88}$  & $235.87\pm79.50$  & $111.92\pm81.49$  & $73.25\pm48.96$   \\
                                                                              & 2 (20 env steps)                                                           & $228.34\pm114.93$ & $206.81\pm121.11$ & $210.17\pm126.19$ & $209.87\pm110.78$ & $117.26\pm87.03$  & $74.12\pm64.32$   \\
                                                                              & 3 (30 env steps)                                                           & $208.74\pm122.31$ & $242.38\pm106.81$ & $250.37\pm91.48$  & $182.82\pm126.36$ & $85.07\pm75.52$   & $76.38\pm60.24$   \\
                                                                              & 5 (50 env steps)                                                           & $209.59\pm115.71$ & $213.64\pm124.66$ & $198.95\pm127.97$ & $149.44\pm124.92$ & $71.60\pm75.53$   & $62.86\pm49.60$   \\ \hline
\multirow{4}{*}{red\_stack\_blue}  & 1 (10 env steps)                                                           & $163.20\pm64.30$  & $\mathbf{169.98\pm61.86}$  & $165.71\pm59.09$  & $137.71\pm54.03$  & $62.64\pm46.53$   & $44.04\pm30.46$   \\
                                                                              & 2 (20 env steps)                                                           & $153.61\pm72.96$  & $142.30\pm80.31$  & $130.36\pm72.65$  & $124.08\pm77.25$  & $69.16\pm54.87$   & $47.70\pm35.69$   \\
                                                                              & 3 (30 env steps)                                                           & $139.09\pm86.27$  & $146.60\pm76.52$  & $116.95\pm87.39$  & $129.47\pm75.73$  & $53.05\pm48.60$   & $47.65\pm34.57$   \\
                                                                              & 5 (50 env steps)                                                           & $134.90\pm77.47$  & $146.47\pm89.70$  & $127.17\pm78.31$  & $104.15\pm83.95$  & $48.57\pm48.87$   & $34.84\pm36.28$   \\ \hline
\multirow{4}{*}{red\_hover\_green} & 1 (10 env steps)                                                           & $252.92\pm83.79$  & $\mathbf{266.28\pm79.53}$  & $242.92\pm92.43$  & $201.36\pm102.38$ & $92.81\pm80.08$   & $82.40\pm65.66$   \\
                                                                              & 2 (20 env steps)                                                           & $253.96\pm104.24$ & $251.61\pm113.12$ & $206.36\pm128.29$ & $193.89\pm123.43$ & $111.49\pm92.15$  & $74.11\pm64.53$   \\
                                                                              & 3 (30 env steps)                                                           & $245.03\pm98.20$  & $257.20\pm94.11$  & $206.95\pm125.21$ & $198.74\pm112.13$ & $102.04\pm83.74$  & $80.75\pm60.20$   \\
                                                                              & 5 (50 env steps)                                                           & $209.99\pm132.24$ & $230.03\pm121.60$ & $151.79\pm118.87$ & $162.21\pm121.88$ & $101.46\pm92.07$  & $63.19\pm60.46$   \\ \hline
\multirow{4}{*}{red\_stack\_green} & 1 (10 env steps)                                                           & $149.23\pm49.02$  & $147.83\pm50.09$  & $147.16\pm63.95$  & $116.38\pm70.68$  & $55.65\pm45.58$   & $53.20\pm36.29$   \\
                                                                              & 2 (20 env steps)                                                           & $\mathbf{162.75\pm73.69}$  & $134.26\pm75.28$  & $143.21\pm63.79$  & $118.14\pm69.70$  & $64.20\pm58.32$   & $34.47\pm32.61$   \\
                                                                              & 3 (30 env steps)                                                           & $148.30\pm70.34$  & $148.54\pm71.91$  & $137.68\pm77.25$  & $130.23\pm74.53$  & $49.89\pm45.62$   & $45.57\pm38.98$   \\
                                                                              & 5 (50 env steps)                                                           & $140.50\pm77.52$  & $141.49\pm71.50$  & $131.12\pm79.70$  & $111.28\pm75.29$  & $50.86\pm46.53$   & $37.15\pm36.43$   \\ \hline
\multirow{4}{*}{bring\_red}                                                   & 1 (10 env steps)                                                           & $200.75\pm73.16$  & $212.23\pm79.63$  & $\mathbf{225.97\pm83.76}$  & $186.15\pm90.86$  & $86.34\pm58.72$   & $75.23\pm44.01$   \\
                                                                              & 2 (20 env steps)                                                           & $203.98\pm97.90$  & $199.17\pm95.42$  & $209.04\pm106.02$ & $172.84\pm92.91$  & $87.21\pm70.99$   & $87.81\pm54.11$   \\
                                                                              & 3 (30 env steps)                                                           & $205.18\pm85.67$  & $194.97\pm109.51$ & $211.89\pm82.38$  & $174.25\pm91.88$  & $82.62\pm65.49$   & $85.13\pm52.48$   \\
                                                                              & 5 (50 env steps)                                                           & $171.37\pm106.52$ & $191.88\pm102.87$ & $189.46\pm85.13$  & $160.37\pm92.36$  & $80.58\pm66.44$   & $62.12\pm52.43$   \\ \hline
\multirow{4}{*}{reach\_green}                                                 & 1 (10 env steps)                                                           & $227.89\pm112.03$ & $\mathbf{241.41\pm104.34}$ & $214.59\pm101.32$ & $202.14\pm85.68$  & $109.02\pm80.77$  & $92.83\pm79.68$   \\
                                                                              & 2 (20 env steps)                                                           & $193.12\pm138.08$ & $195.61\pm118.21$ & $181.52\pm119.60$ & $146.79\pm88.83$  & $92.92\pm74.86$   & $76.14\pm59.45$   \\
                                                                              & 3 (30 env steps)                                                           & $170.56\pm124.03$ & $190.91\pm130.19$ & $155.38\pm109.42$ & $120.62\pm96.02$  & $90.34\pm80.37$   & $76.41\pm63.71$   \\
                                                                              & 5 (50 env steps)                                                           & $121.01\pm114.82$ & $118.40\pm113.99$ & $100.52\pm90.30$  & $94.40\pm86.08$   & $107.70\pm89.71$  & $83.71\pm71.01$   \\ \hline
\multirow{4}{*}{lift\_green}                                                  & 1 (10 env steps)                                                           & $\mathbf{59.75\pm26.74}$   & $59.04\pm26.39$   & $57.56\pm22.99$   & $46.63\pm23.13$   & $24.71\pm13.88$   & $24.06\pm21.63$   \\
                                                                              & 2 (20 env steps)                                                           & $48.25\pm32.03$   & $42.05\pm30.42$   & $48.14\pm28.42$   & $43.49\pm25.45$   & $21.16\pm15.21$   & $17.62\pm14.64$   \\
                                                                              & 3 (30 env steps)                                                           & $31.77\pm28.26$   & $32.66\pm23.80$   & $33.04\pm26.56$   & $35.87\pm27.42$   & $20.48\pm15.45$   & $23.25\pm19.40$   \\
                                                                              & 5 (50 env steps)                                                           & $21.44\pm21.53$   & $29.97\pm26.10$   & $32.77\pm25.60$   & $24.08\pm24.74$   & $20.79\pm20.59$   & $20.67\pm18.09$  
\end{tabular}
    \captionsetup{font=footnotesize,singlelinecheck=off}
\caption{
Additional evaluation results for "Zero-shot via Planning" with context length $K=10$.
The results for each evaluation task are listed by their planning horizon $H$ 
($H$ operates in $z$ and each $z$ corresponds to $K$ primitive action steps).
The "max reward" stands for the
maximum single step reward 
($[0,1]$)
achieved during an evaluation episode,
while the "return" is the undiscounted cumulative reward obtained within an episode
(the maximum number of steps for each episode is $400$).
Each evaluation configuration is conducted with $50$ random seeds.
}
    \label{tab:app-zero-shot}
\end{sidewaystable}

In Table \ref{tab:app-zero-shot},
we present further evaluation results for jumpy planning ("Zero-shot via Planning").
Each task is evaluated 
by varying:
1. the planning horizon $H$
($H=1,2,3,5$ corresponds to looking ahead 10, 20, 30, 50 steps respectively in terms of primitive actions
since the latent horizon $K=10$);
2. the number of primitive actions executed in each planning iteration,
in other words,
once the planning procedure gives out a planned $z$, how many steps it is kept fixed in the LL policy to give out primitive actions.
Each configuration is evaluated with 50 episodes with random seeds,
with a maximum number of 400 steps for each episode.
The performance of the agent on these tasks are descent,
while it generally drops slightly with longer planning horizon,
which could be due to accumulated prediction errors in the unrolling of the jumpy model.
With more number of primitive actions executed per planned latent,
the obtained returns are generally quite consistent for $1,2,10$ actions
(note that the context length $K$ is set to 10 during phase-1),
which shows that the proposed method
not only enables jumpy planning but could also support task accomplishment at lower control rate.
When one latent is held fixed for as long as $100$ or $200$ steps,
the performance drops greatly,
as the latent skill was not trained to be able to provide meaningful context for that many steps
(also note that each episode is at most 400 steps long).

\begin{sidewaystable}
\centering
\footnotesize
\begin{tabular}{cc|ccccccc}
\multicolumn{2}{c|}{Planning horizon}           & 1                & 2                & 3                 & 5                & 10               & 30               & 50               \\ \hline
\multirow{2}{*}{reach\_red}        & Max reward & $1.00\pm0.00$    & $1.00\pm0.00$    & $0.97\pm0.16$     & $0.96\pm0.13$    & $0.95\pm0.17$    & $0.94\pm0.21$    & $0.94\pm0.18$    \\
                                   & Return     & $368.01\pm19.76$ & $\mathbf{375.96\pm17.86}$ & $365.91\pm67.10$  & $352.50\pm77.14$ & $350.40\pm79.13$ & $328.38\pm92.97$ & $316.05\pm96.02$ \\ \hline
\multirow{2}{*}{lift\_red}         & Max reward & $0.65\pm0.30$    & $0.59\pm0.24$    & $0.56\pm0.27$     & $0.30\pm0.18$    & $0.40\pm0.15$    & $0.36\pm0.13$    & $0.35\pm0.14$    \\
                                   & Return     & $\mathbf{130.16\pm52.55}$ & $127.66\pm51.96$ & $120.80\pm62.28$  & $104.65\pm35.38$ & $89.86\pm20.78$  & $86.37\pm14.32$  & $80.07\pm17.74$  \\ \hline
\multirow{2}{*}{red\_hover\_blue}  & Max reward & $0.51\pm0.30$    & $0.56\pm0.34$    & $0.60\pm0.29$     & $0.47\pm0.29$    & $0.36\pm0.23$    & $0.37\pm0.28$    & $0.31\pm0.25$    \\
                                   & Return     & $71.54\pm18.72$  & $73.26\pm37.90$  & $\mathbf{73.41\pm20.02}$   & $66.62\pm19.20$  & $64.20\pm23.41$  & $56.84\pm21.34$  & $47.55\pm18.66$  \\ \hline
\multirow{2}{*}{red\_stack\_blue}  & Max reward & $0.37\pm0.21$    & $0.41\pm0.23$    & $0.43\pm0.25$     & $0.29\pm0.18$    & $0.27\pm0.20$    & $0.22\pm0.15$    & $0.21\pm0.17$    \\
                                   & Return     & $47.98\pm25.23$  & $50.20\pm22.62$  & $\mathbf{57.28\pm33.95}$   & $40.30\pm15.03$  & $42.19\pm28.93$  & $35.05\pm7.30$   & $29.11\pm10.59$  \\ \hline
\multirow{2}{*}{red\_hover\_green} & Max reward & $0.51\pm0.29$    & $0.67\pm0.30$    & $0.63\pm0.31$     & $0.49\pm0.28$    & $0.38\pm0.24$    & $0.37\pm0.27$    & $0.30\pm0.24$    \\
                                   & Return     & $69.30\pm11.82$  & $\mathbf{84.78\pm40.51}$  & $79.34\pm29.36$   & $68.91\pm15.37$  & $61.49\pm13.72$  & $55.74\pm16.47$  & $44.65\pm19.34$  \\ \hline
\multirow{2}{*}{red\_stack\_green} & Max reward & $0.35\pm0.17$    & $0.39\pm0.18$    & $0.33\pm0.19$     & $0.29\pm0.18$    & $0.21\pm0.16$    & $0.20\pm0.13$    & $0.22\pm0.17$    \\
                                   & Return     & $44.03\pm13.06$  & $\mathbf{46.09\pm13.43}$  & $41.15\pm15.61$   & $42.20\pm21.56$  & $37.07\pm10.11$  & $34.84\pm8.08$   & $29.24\pm12.21$  \\ \hline
\multirow{2}{*}{bring\_red}        & Max reward & $0.64\pm0.33$    & $0.59\pm0.32$    & $0.64\pm0.34$     & $0.53\pm0.32$    & $0.28\pm0.19$    & $0.15\pm0.15$    & $0.16\pm0.15$    \\
                                   & Return     & $121.06\pm62.10$ & $113.10\pm70.74$ & $\mathbf{132.52\pm100.68}$ & $106.16\pm86.65$ & $55.37\pm31.69$  & $26.14\pm28.51$  & $31.86\pm31.88$  \\ \hline
\multirow{2}{*}{reach\_green}      & Max reward & $1.00\pm0.00$    & $0.94\pm0.23$    & $1.00\pm0.00$     & $0.94\pm0.22$    & $0.95\pm0.20$    & $0.98\pm0.13$    & $0.96\pm0.17$    \\
                                   & Return     & $\mathbf{375.54\pm16.06}$ & $356.82\pm90.78$ & $371.84\pm42.32$  & $354.62\pm89.26$ & $350.90\pm78.94$ & $346.39\pm57.33$ & $332.16\pm76.69$ \\ \hline
\multirow{2}{*}{lift\_green}       & Max reward & $0.50\pm0.27$    & $0.54\pm0.23$    & $0.54\pm0.22$     & $0.43\pm0.16$    & $0.37\pm0.15$    & $0.36\pm0.15$    & $0.30\pm0.12$    \\
                                   & Return     & $108.53\pm36.17$ & $\mathbf{114.77\pm46.45}$ & $105.76\pm42.03$  & $91.84\pm19.21$  & $87.14\pm23.71$  & $86.00\pm10.32$  & $77.38\pm20.72$ 
\end{tabular}
\captionsetup{font=footnotesize,singlelinecheck=off}
\caption{
Additional evaluation results for "Zero-shot via Planning (K=1)".
The results for each evaluation task are listed by their planning horizon 
(in terms of primitive action steps).
The "max reward" stands for the
maximum single step reward 
($[0,1]$)
achieved during an evaluation episode,
while the "return" is the undiscounted cumulative reward obtained within an episode
(the maximum number of steps for each episode is $400$).
Each evaluation configuration is conducted with $50$ random seeds.
}
\label{tab:app-zero-shot-1step}
\end{sidewaystable}

In Table \ref{tab:app-zero-shot-1step},
we present more detailed zero-shot evaluation results for non-jumpy planning ("Zero-shot via Planning ($K=1$)").
Each task is evaluated with planning horizon $H=1,2,3,5,10,30,50$
looking ahead steps in terms of primitive actions.
Each configuration is evaluated with 50 episodes with random seeds,
with a maximum number of steps of 400 for each episode.
The performance of the agent on simpler tasks like reach is quite decent,
even on out-of-distribution task like reach\_green,
probably due to the fact that one step transition dynamics is universal across different task in the same environment and is sufficient for guiding short horizon tasks.
While its performance on all other more complicated tasks drop substantially compared to the one trained with $K=10$,
this should be due to that for tackling long horizon tasks,
shorter planning horizon does not suffice while bigger look ahead steps lead to higher accumulation error with one-step models.

\begin{figure}[b]
    \centering
    \begin{subfigure}{0.13\textwidth}
        \centering
        \includegraphics[width=\linewidth]{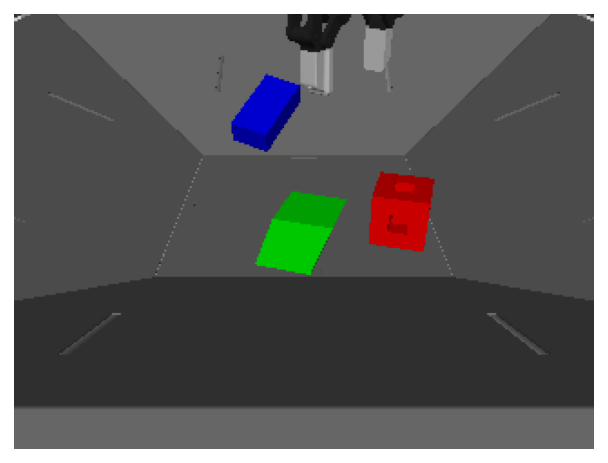}
    \end{subfigure}
    \begin{subfigure}{0.13\textwidth}
        \centering
        \includegraphics[width=\linewidth]{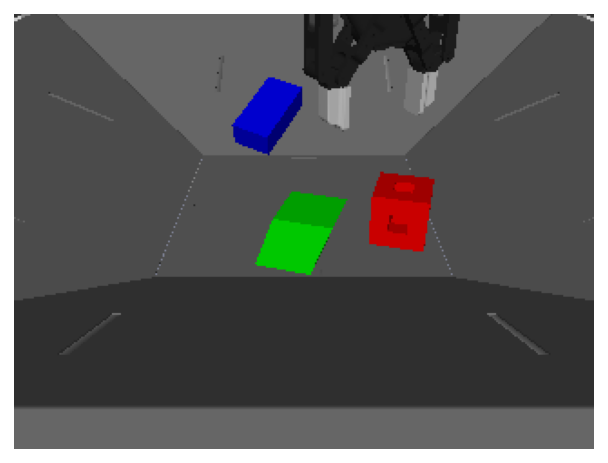}
    \end{subfigure}
    \begin{subfigure}{0.13\textwidth}
        \centering
        \includegraphics[width=\linewidth]{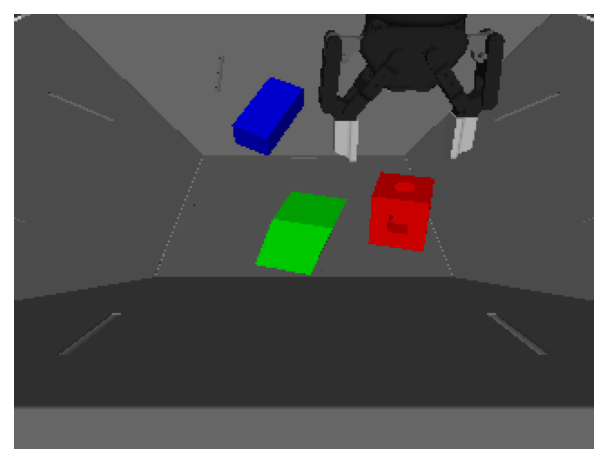}
    \end{subfigure}
    \begin{subfigure}{0.13\textwidth}
        \centering
        \includegraphics[width=\linewidth]{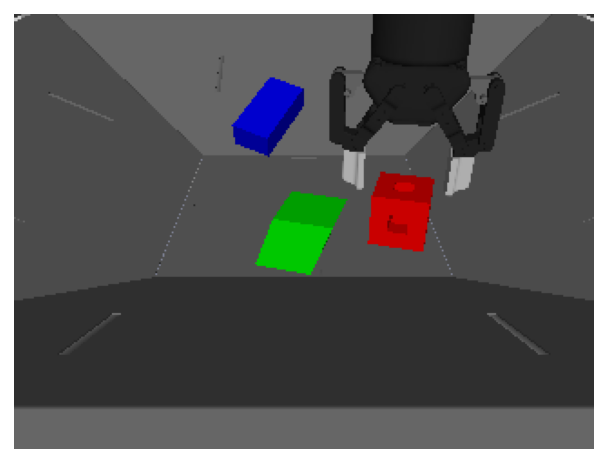}
    \end{subfigure}
    \begin{subfigure}{0.13\textwidth}
        \centering
        \includegraphics[width=\linewidth]{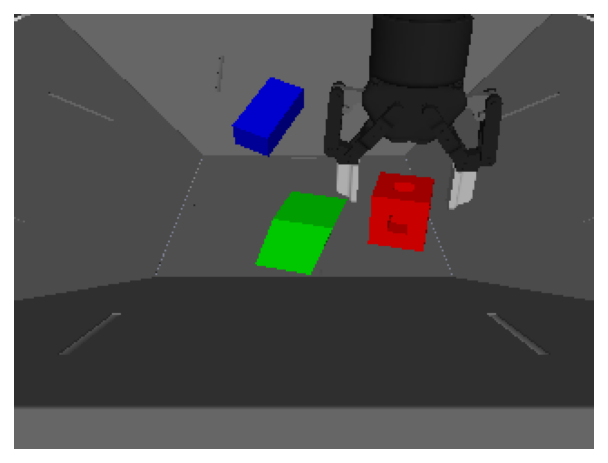}
    \end{subfigure}
    \begin{subfigure}{0.13\textwidth}
        \centering
        \includegraphics[width=\linewidth]{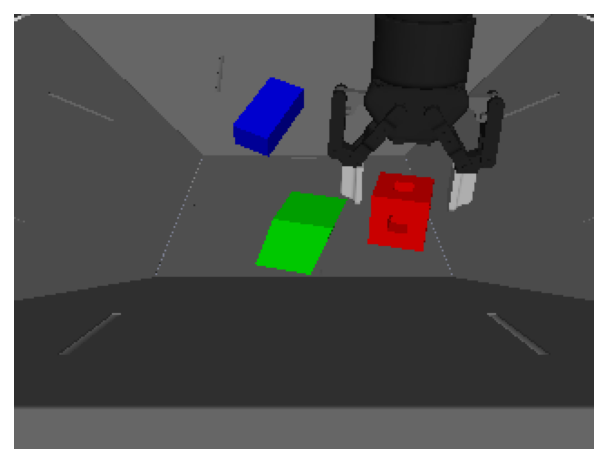}
    \end{subfigure}
    \begin{subfigure}{0.13\textwidth}
        \centering
        \includegraphics[width=\linewidth]{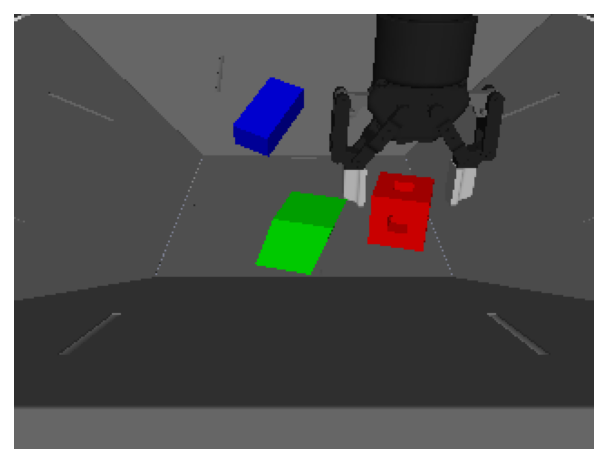}
    \end{subfigure}

    ~

    \centering
    \begin{subfigure}{0.13\textwidth}
        \centering
        \includegraphics[width=\linewidth]{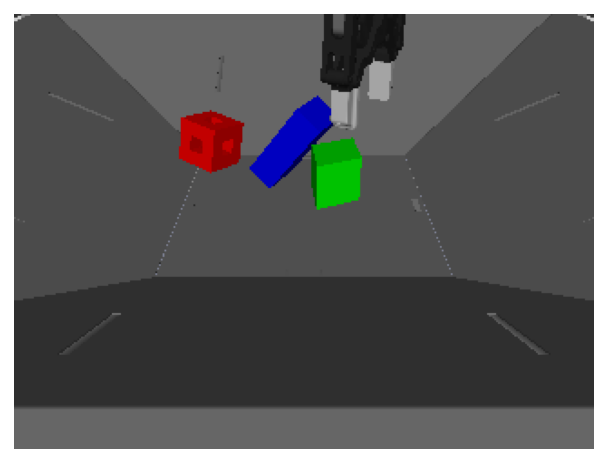}
    \end{subfigure}
    \begin{subfigure}{0.13\textwidth}
        \centering
        \includegraphics[width=\linewidth]{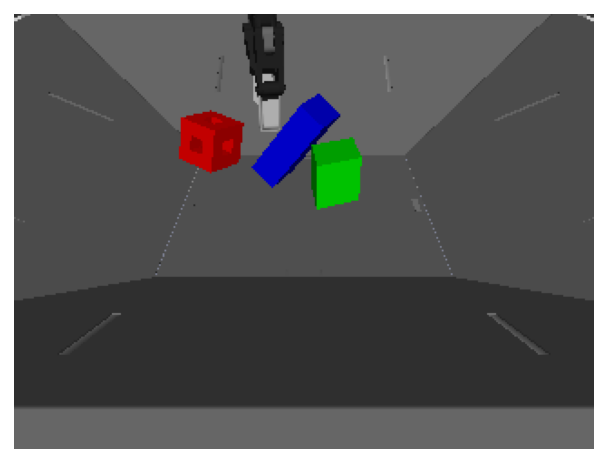}
    \end{subfigure}
    \begin{subfigure}{0.13\textwidth}
        \centering
        \includegraphics[width=\linewidth]{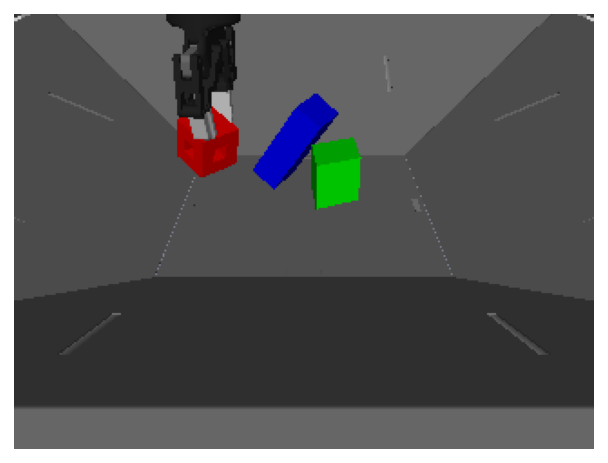}
    \end{subfigure}
    \begin{subfigure}{0.13\textwidth}
        \centering
        \includegraphics[width=\linewidth]{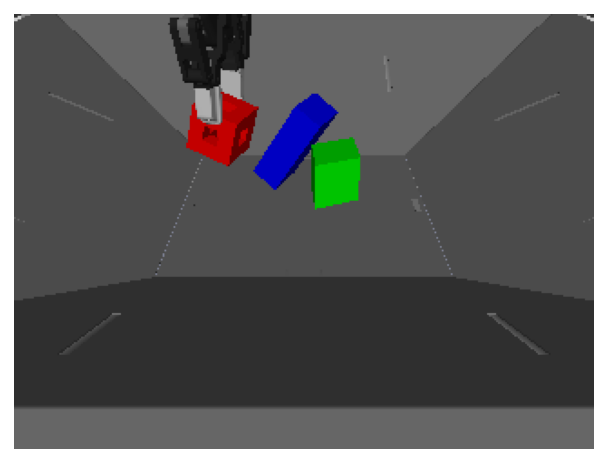}
    \end{subfigure}
    \begin{subfigure}{0.13\textwidth}
        \centering
        \includegraphics[width=\linewidth]{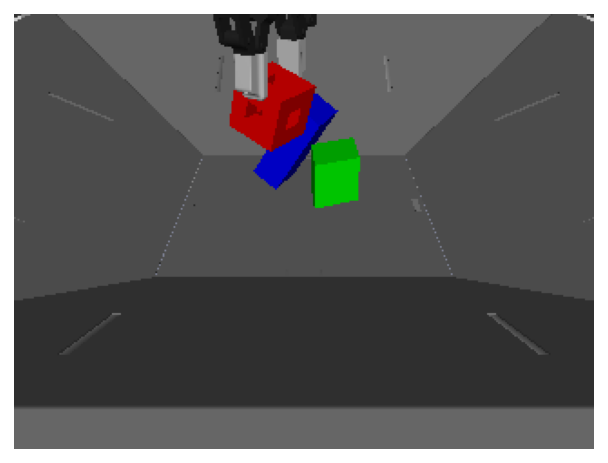}
    \end{subfigure}
    \begin{subfigure}{0.13\textwidth}
        \centering
        \includegraphics[width=\linewidth]{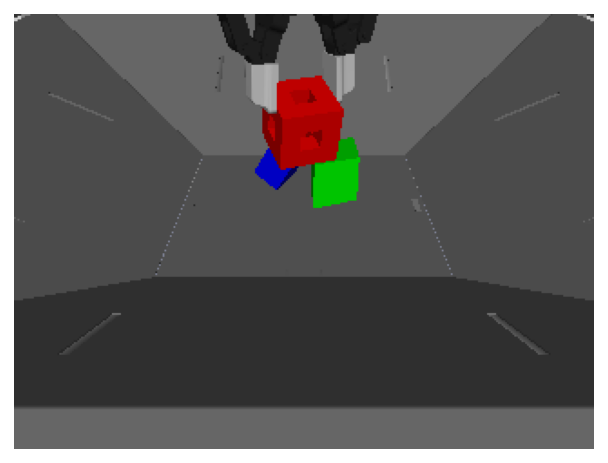}
    \end{subfigure}
    \begin{subfigure}{0.13\textwidth}
        \centering
        \includegraphics[width=\linewidth]{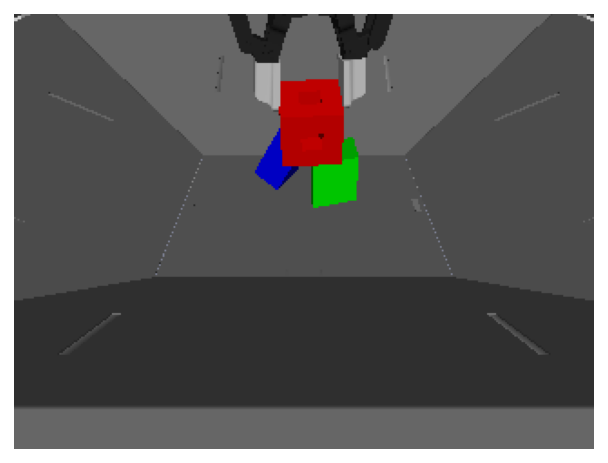}
    \end{subfigure}

    ~
    
    \centering
    \begin{subfigure}{0.13\textwidth}
        \centering
        \includegraphics[width=\linewidth]{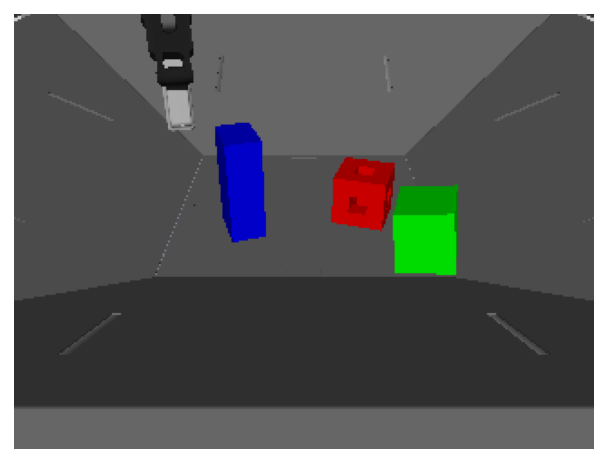}
    \end{subfigure}
    \begin{subfigure}{0.13\textwidth}
        \centering
        \includegraphics[width=\linewidth]{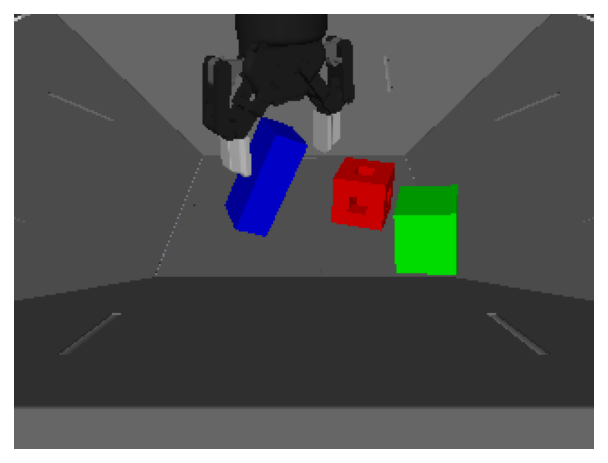}
    \end{subfigure}
    \begin{subfigure}{0.13\textwidth}
        \centering
        \includegraphics[width=\linewidth]{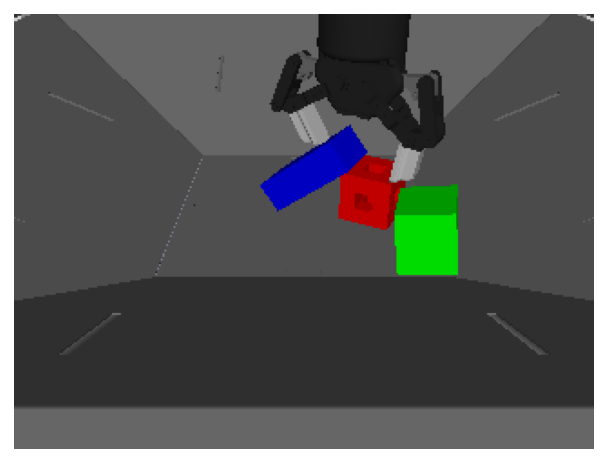}
    \end{subfigure}
    \begin{subfigure}{0.13\textwidth}
        \centering
        \includegraphics[width=\linewidth]{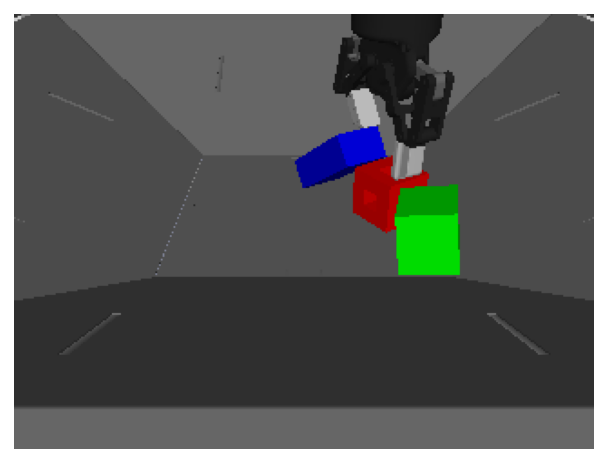}
    \end{subfigure}
    \begin{subfigure}{0.13\textwidth}
        \centering
        \includegraphics[width=\linewidth]{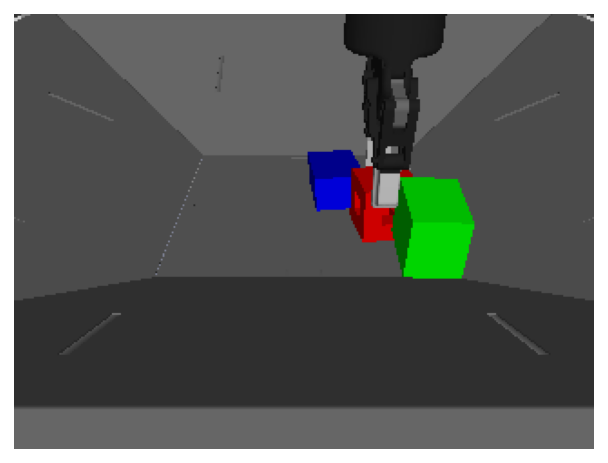}
    \end{subfigure}
    \begin{subfigure}{0.13\textwidth}
        \centering
        \includegraphics[width=\linewidth]{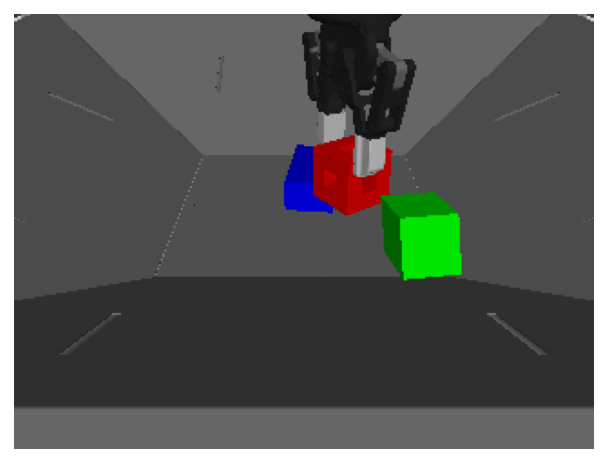}
    \end{subfigure}
    \begin{subfigure}{0.13\textwidth}
        \centering
        \includegraphics[width=\linewidth]{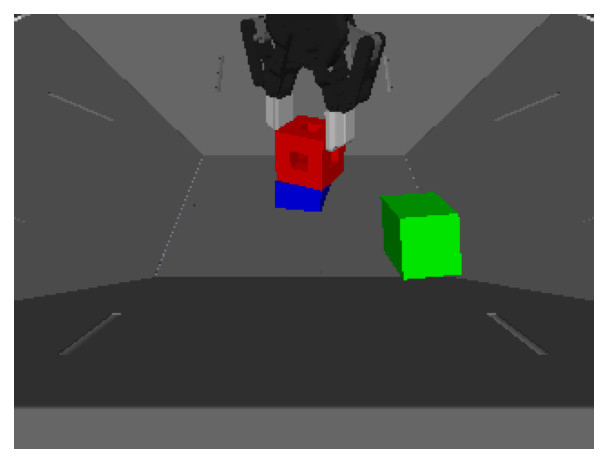}
    \end{subfigure}

    ~

    \centering
    \begin{subfigure}{0.13\textwidth}
        \centering
        \includegraphics[width=\linewidth]{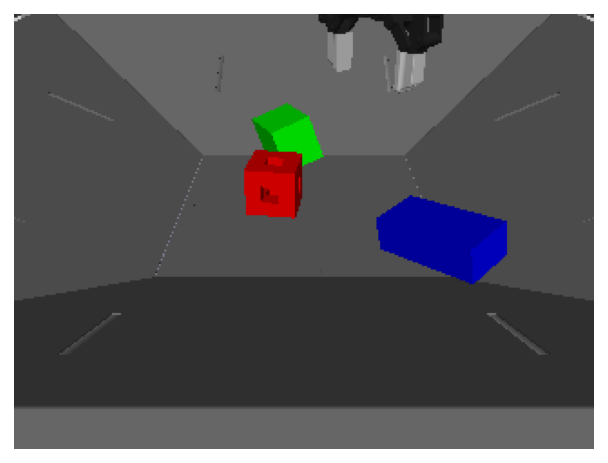}
    \end{subfigure}
    \begin{subfigure}{0.13\textwidth}
        \centering
        \includegraphics[width=\linewidth]{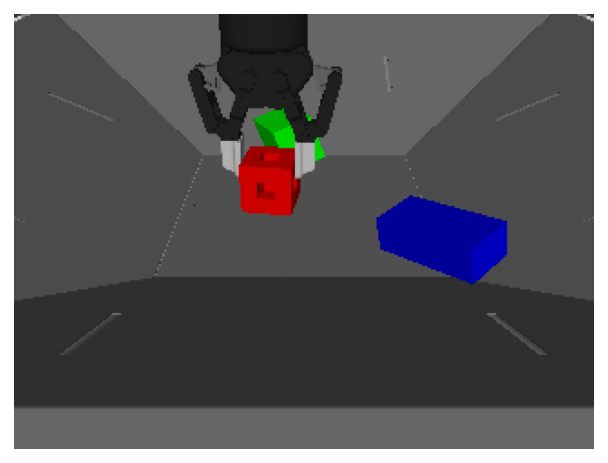}
    \end{subfigure}
    \begin{subfigure}{0.13\textwidth}
        \centering
        \includegraphics[width=\linewidth]{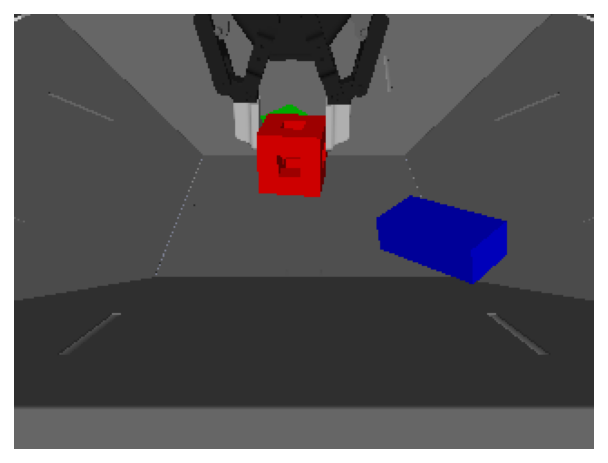}
    \end{subfigure}
    \begin{subfigure}{0.13\textwidth}
        \centering
        \includegraphics[width=\linewidth]{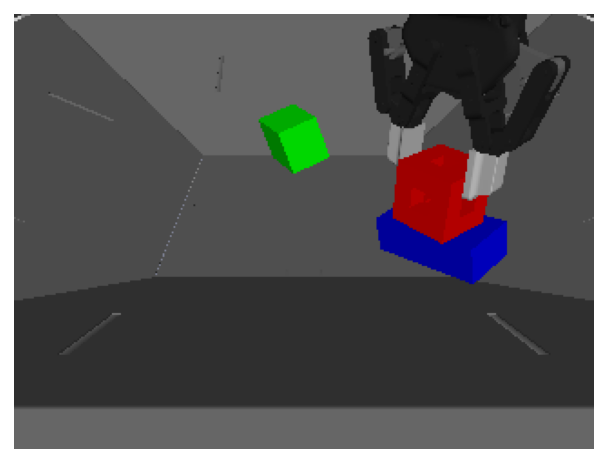}
    \end{subfigure}
    \begin{subfigure}{0.13\textwidth}
        \centering
        \includegraphics[width=\linewidth]{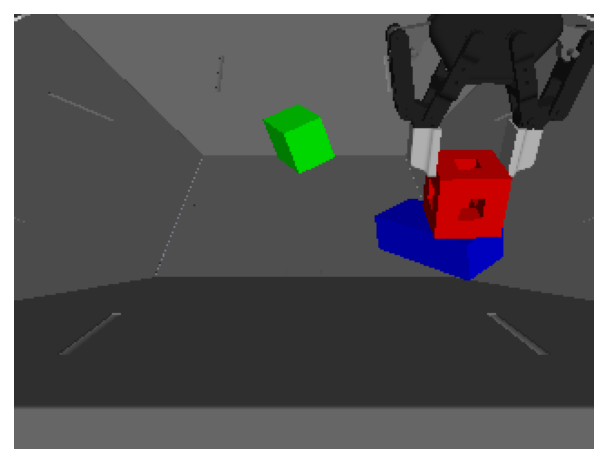}
    \end{subfigure}
    \begin{subfigure}{0.13\textwidth}
        \centering
        \includegraphics[width=\linewidth]{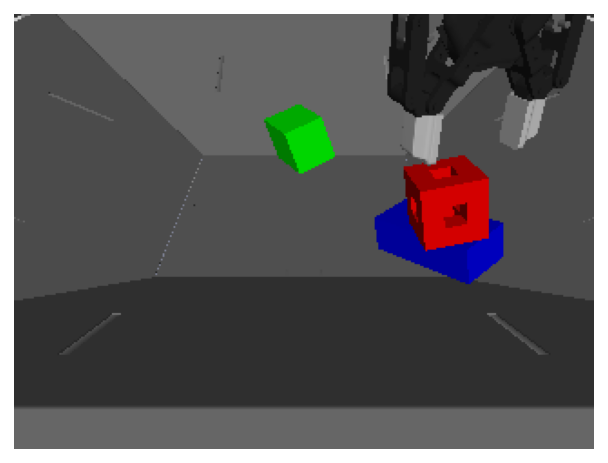}
    \end{subfigure}
    \begin{subfigure}{0.13\textwidth}
        \centering
        \includegraphics[width=\linewidth]{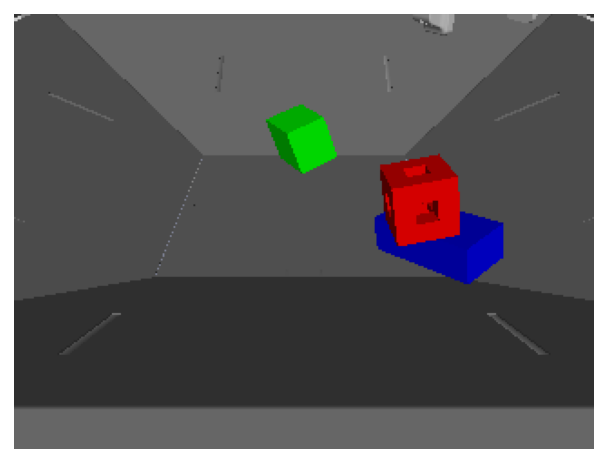}
    \end{subfigure}

    ~
    
    \centering
    \begin{subfigure}{0.13\textwidth}
        \centering
        \includegraphics[width=\linewidth]{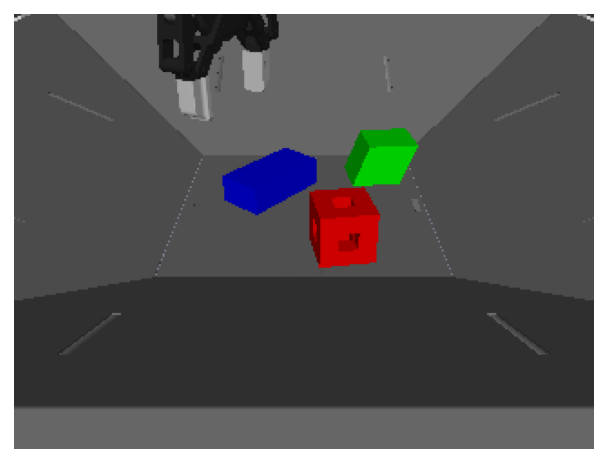}
    \end{subfigure}
    \begin{subfigure}{0.13\textwidth}
        \centering
        \includegraphics[width=\linewidth]{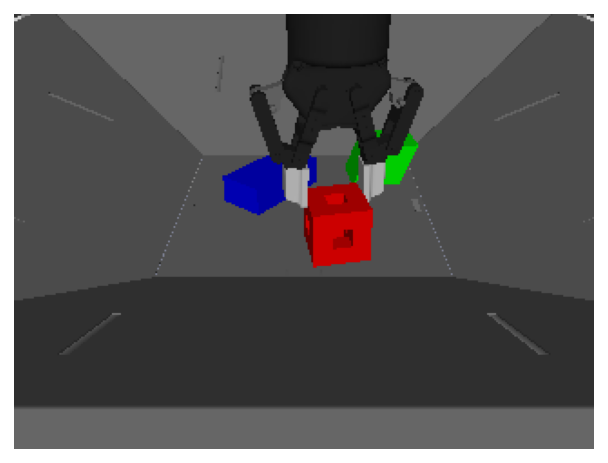}
    \end{subfigure}
    \begin{subfigure}{0.13\textwidth}
        \centering
        \includegraphics[width=\linewidth]{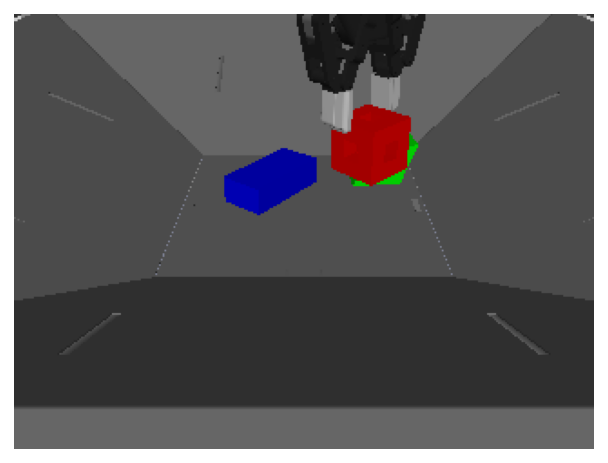}
    \end{subfigure}
    \begin{subfigure}{0.13\textwidth}
        \centering
        \includegraphics[width=\linewidth]{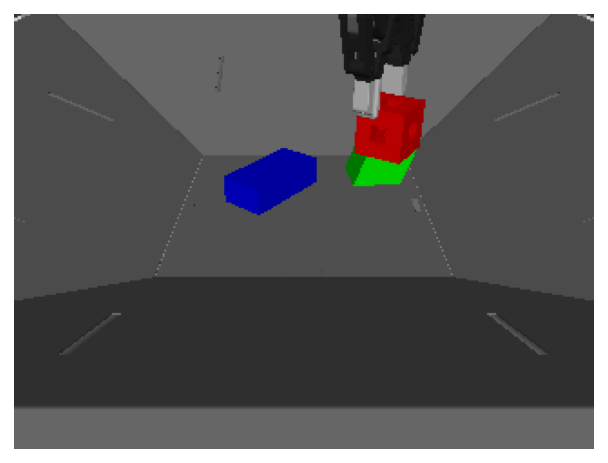}
    \end{subfigure}
    \begin{subfigure}{0.13\textwidth}
        \centering
        \includegraphics[width=\linewidth]{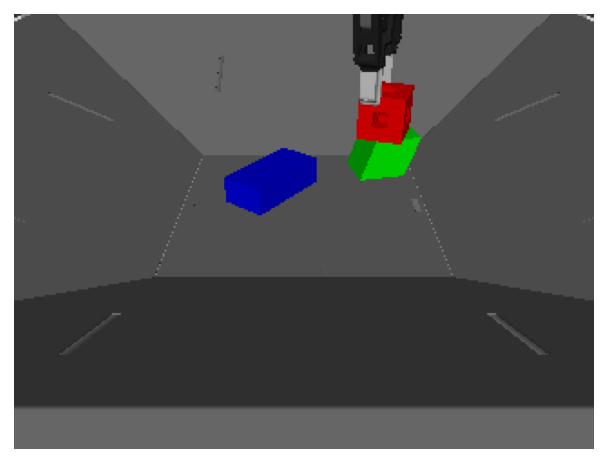}
    \end{subfigure}
    \begin{subfigure}{0.13\textwidth}
        \centering
        \includegraphics[width=\linewidth]{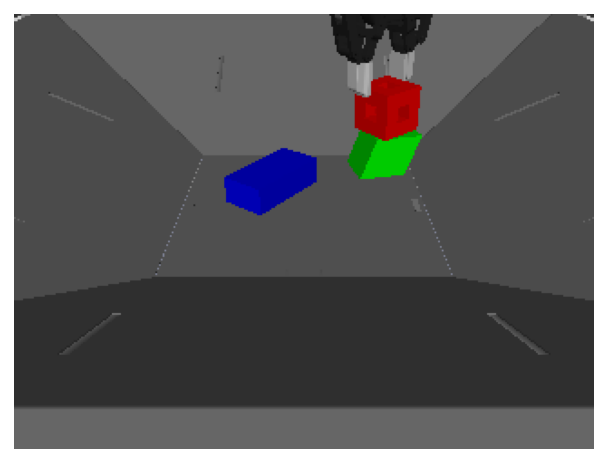}
    \end{subfigure}
    \begin{subfigure}{0.13\textwidth}
        \centering
        \includegraphics[width=\linewidth]{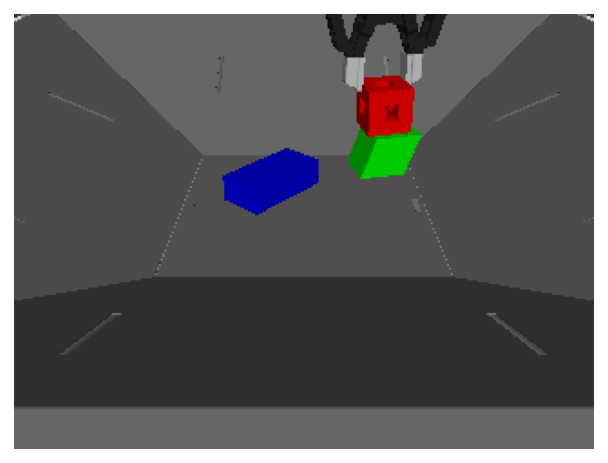}
    \end{subfigure}

    ~
    
    \centering
    \begin{subfigure}{0.13\textwidth}
        \centering
        \includegraphics[width=\linewidth]{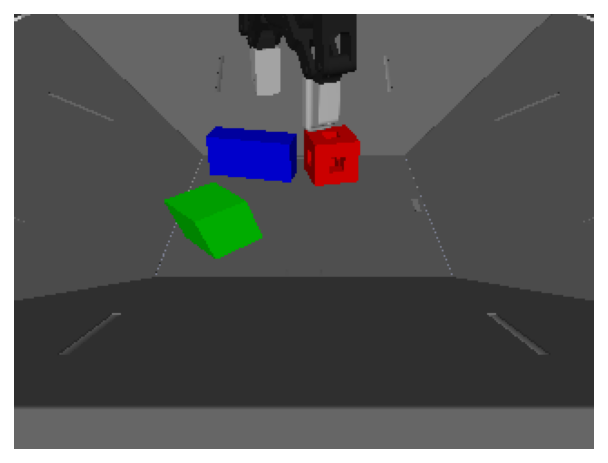}
    \end{subfigure}
    \begin{subfigure}{0.13\textwidth}
        \centering
        \includegraphics[width=\linewidth]{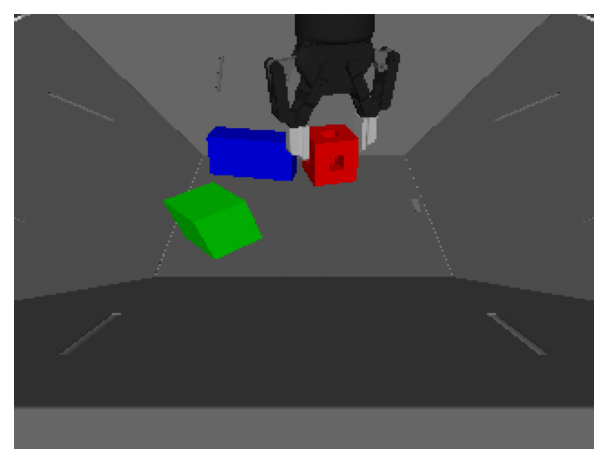}
    \end{subfigure}
    \begin{subfigure}{0.13\textwidth}
        \centering
        \includegraphics[width=\linewidth]{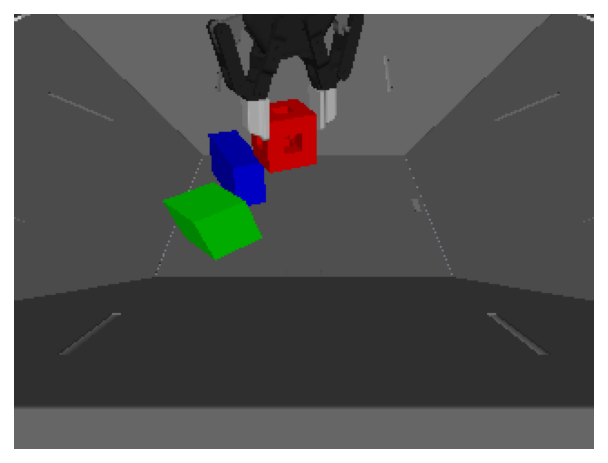}
    \end{subfigure}
    \begin{subfigure}{0.13\textwidth}
        \centering
        \includegraphics[width=\linewidth]{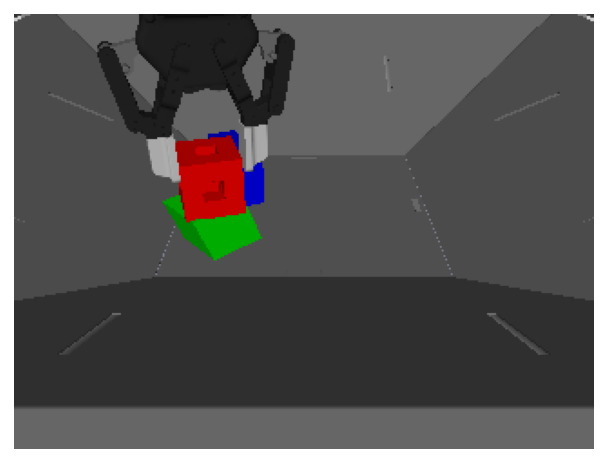}
    \end{subfigure}
    \begin{subfigure}{0.13\textwidth}
        \centering
        \includegraphics[width=\linewidth]{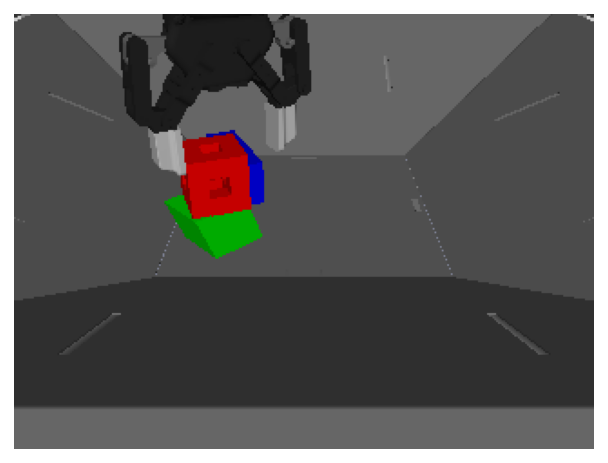}
    \end{subfigure}
    \begin{subfigure}{0.13\textwidth}
        \centering
        \includegraphics[width=\linewidth]{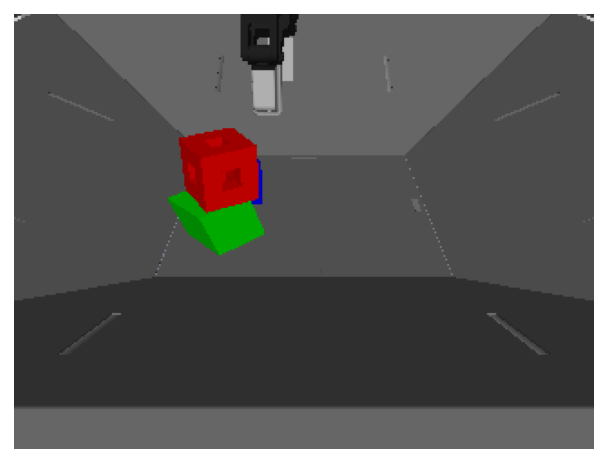}
    \end{subfigure}
    \begin{subfigure}{0.13\textwidth}
        \centering
        \includegraphics[width=\linewidth]{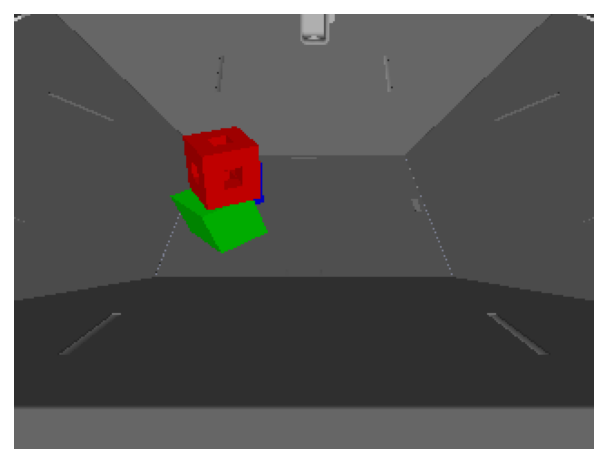}
    \end{subfigure}

    ~

    \centering
    \begin{subfigure}{0.13\textwidth}
        \centering
        \includegraphics[width=\linewidth]{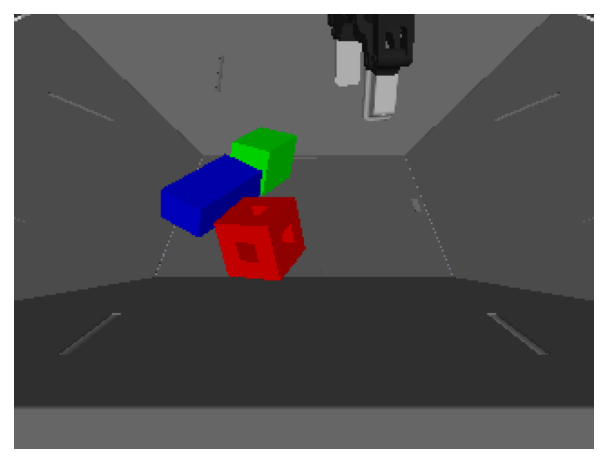}
    \end{subfigure}
    \begin{subfigure}{0.13\textwidth}
        \centering
        \includegraphics[width=\linewidth]{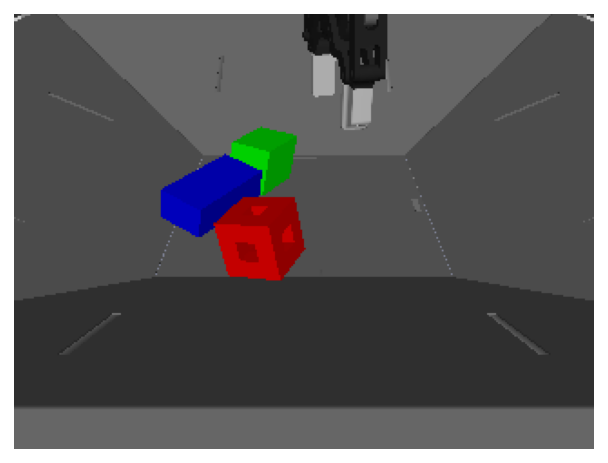}
    \end{subfigure}
    \begin{subfigure}{0.13\textwidth}
        \centering
        \includegraphics[width=\linewidth]{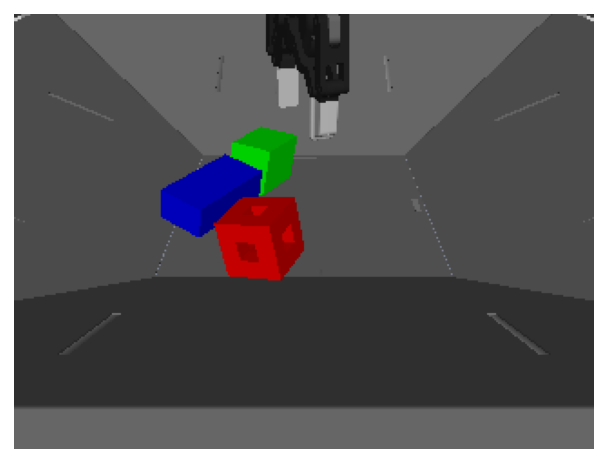}
    \end{subfigure}
    \begin{subfigure}{0.13\textwidth}
        \centering
        \includegraphics[width=\linewidth]{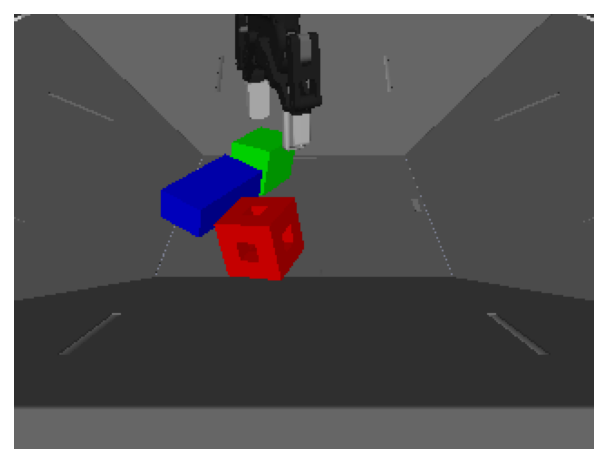}
    \end{subfigure}
    \begin{subfigure}{0.13\textwidth}
        \centering
        \includegraphics[width=\linewidth]{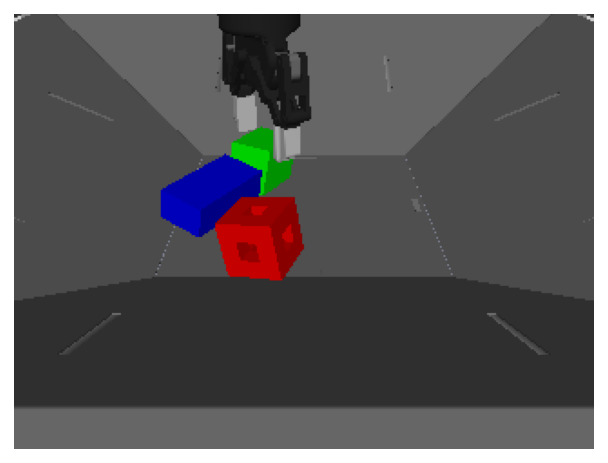}
    \end{subfigure}
    \begin{subfigure}{0.13\textwidth}
        \centering
        \includegraphics[width=\linewidth]{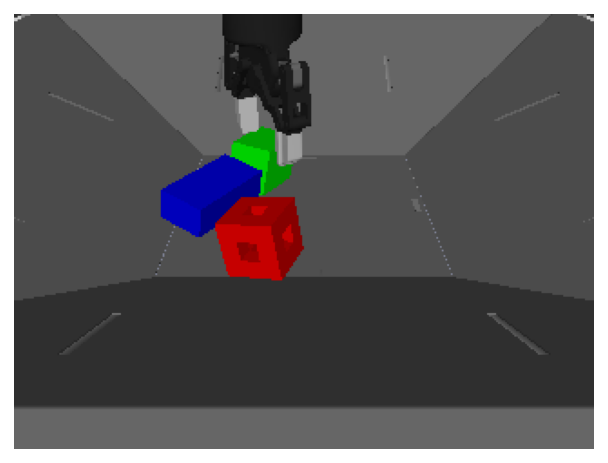}
    \end{subfigure}
    \begin{subfigure}{0.13\textwidth}
        \centering
        \includegraphics[width=\linewidth]{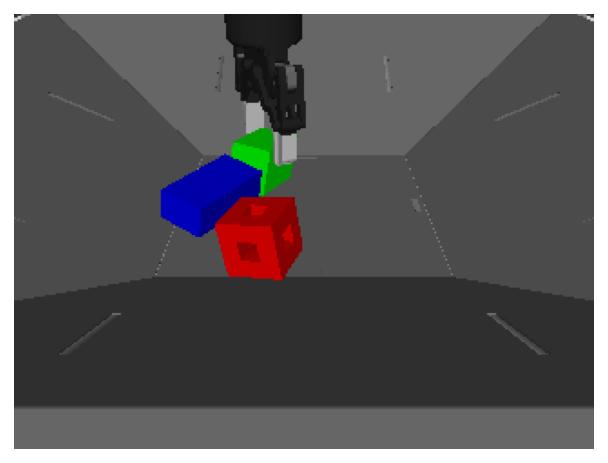}
    \end{subfigure}

    ~
    
    \centering
    \begin{subfigure}{0.13\textwidth}
        \centering
        \includegraphics[width=\linewidth]{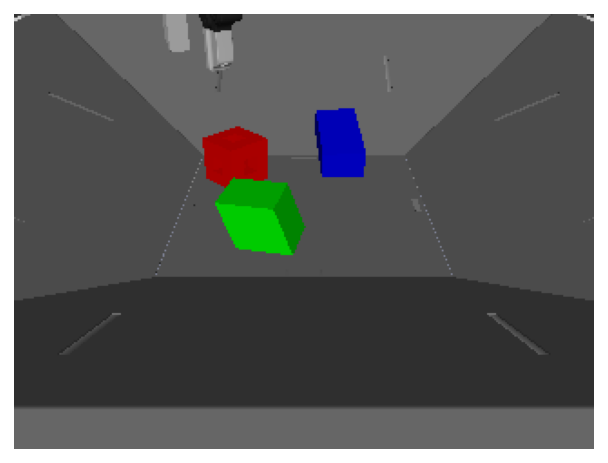}
    \end{subfigure}
    \begin{subfigure}{0.13\textwidth}
        \centering
        \includegraphics[width=\linewidth]{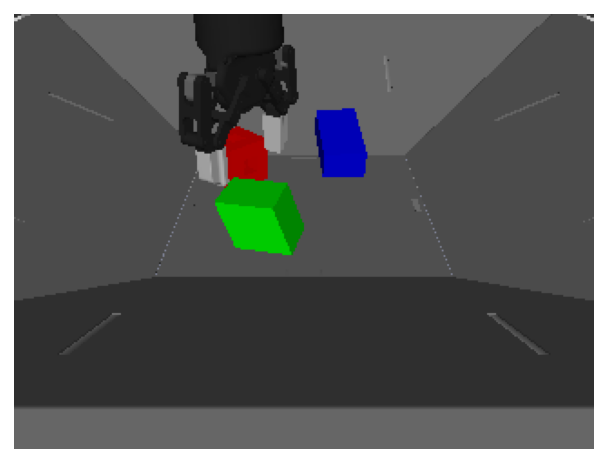}
    \end{subfigure}
    \begin{subfigure}{0.13\textwidth}
        \centering
        \includegraphics[width=\linewidth]{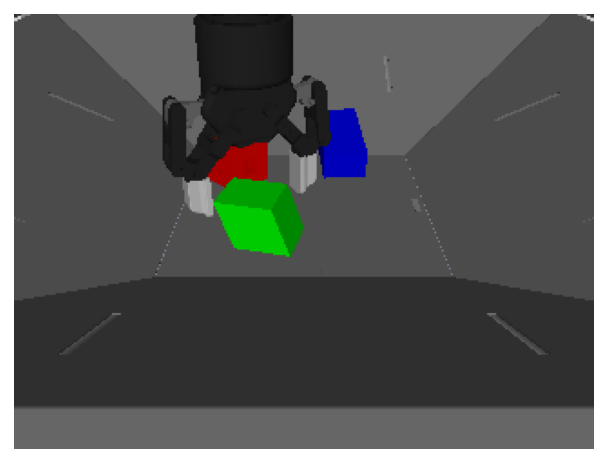}
    \end{subfigure}
    \begin{subfigure}{0.13\textwidth}
        \centering
        \includegraphics[width=\linewidth]{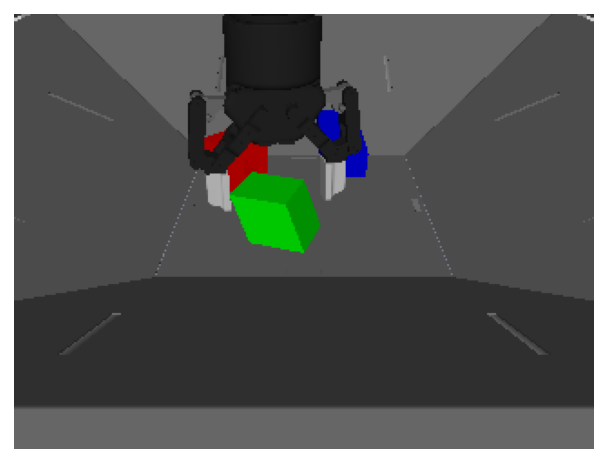}
    \end{subfigure}
    \begin{subfigure}{0.13\textwidth}
        \centering
        \includegraphics[width=\linewidth]{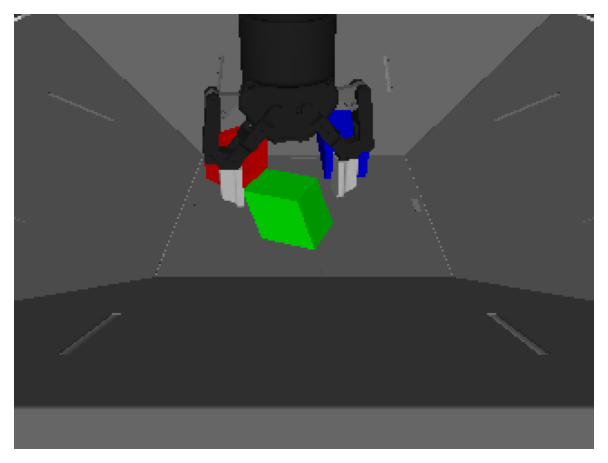}
    \end{subfigure}
    \begin{subfigure}{0.13\textwidth}
        \centering
        \includegraphics[width=\linewidth]{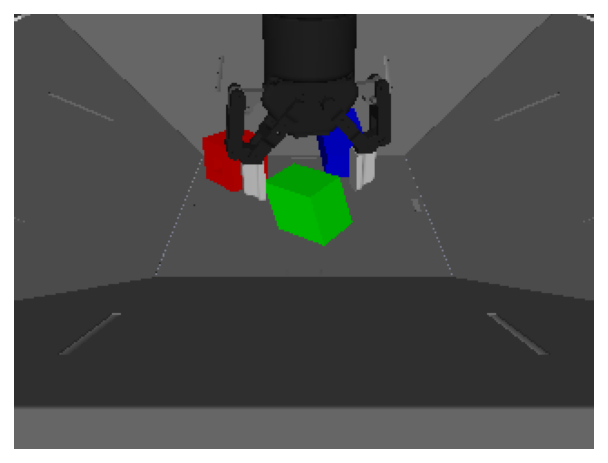}
    \end{subfigure}
    \begin{subfigure}{0.13\textwidth}
        \centering
        \includegraphics[width=\linewidth]{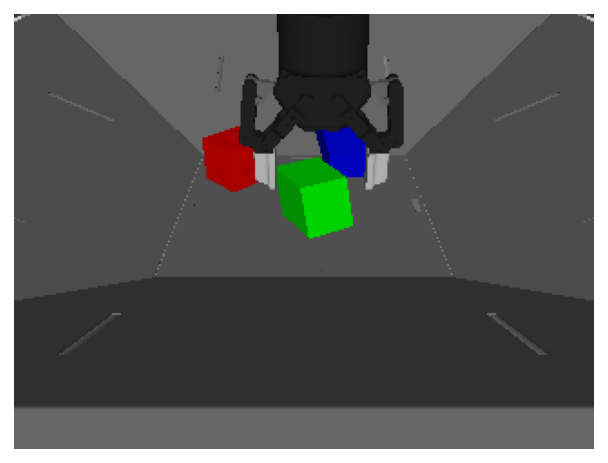}
    \end{subfigure}
   
    \caption{
    Example trajectories of "RL-HL + Planning" on tasks 
    (top to bottom): 
    reach\_red, 
    lift\_red, 
    red\_hover\_blue, 
    red\_stack\_blue, 
    red\_hover\_green, 
    red\_stack\_green, 
    reach\_green and
    lift\_green.
    }
    \label{fig:trajs}
\end{figure}

In Figure \ref{fig:trajs},
we show exemplar trajectories of "RL-HL + Planning" for all evaluated tasks to clarity the tasks considered.

\end{document}